\documentclass{article}

     \usepackage[nonatbib,final]{neurips_2023}

\usepackage[utf8]{inputenc} %
\usepackage[T1]{fontenc}    %
\usepackage{hyperref}       %
\usepackage{url}            %
\usepackage{booktabs}       %
\usepackage{amsfonts}       %
\usepackage{nicefrac}       %
\usepackage{microtype}      %
\usepackage{epsfig}
\usepackage{graphicx}
\usepackage{tipa} %
\usepackage{textcomp} %
\usepackage{MnSymbol} %
\usepackage{xspace}
\usepackage{tabularx}
\usepackage[export]{adjustbox} %
\usepackage[normalem]{ulem}

\usepackage[dvipsnames]{xcolor}

\newcommand{\THETITLE}{Kiki or Bouba? Sound Symbolism in Vision-and-Language Models}

\newbox\jsavebox
\newcommand{\jsubfig}[2]{%
	\sbox\jsavebox{#1}%
	\parbox[t]{\wd\jsavebox}{\centering\usebox\jsavebox\\#2}%
	}

\newcommand{\phon}[1]{\texttt{/\textipa{#1}/}}
\newcommand{\graphemic}[1]{\texttt{\textlangle#1\textrangle }}

\newcommand{\ROUND}{$\bigcircle$\xspace}
\newcommand{\SHARP}{$\largestar$\xspace}
\newcommand{\NEUTRAL}{--\xspace}

\newcommand{\CR}{C\textsubscript{\ROUND}\xspace}
\newcommand{\CS}{C\textsubscript{\SHARP}\xspace}
\newcommand{\VR}{V\textsubscript{\ROUND}\xspace}
\newcommand{\VS}{V\textsubscript{\SHARP}\xspace}
\newcommand{\VN}{V\textsubscript{\NEUTRAL}\xspace}

\newcommand{\PseudoR}{$\Psi$\textsubscript{\ROUND}\xspace}
\newcommand{\PseudoS}{$\Psi$\textsubscript{\SHARP}\xspace}
\newcommand{\Pseudo}{$\Psi$\xspace}
\newcommand{\Adj}{A\xspace}
\newcommand{\AdjR}{A\textsubscript{\ROUND}\xspace}
\newcommand{\AdjS}{A\textsubscript{\SHARP}\xspace}
\newcommand{\Noun}{N\xspace}

\newcommand{\CRnospace}{C\textsubscript{\ROUND}}
\newcommand{\CSnospace}{C\textsubscript{\SHARP}}
\newcommand{\VRnospace}{V\textsubscript{\ROUND}}
\newcommand{\VSnospace}{V\textsubscript{\SHARP}}

\newcommand{\PseudoSnospace}{$\Psi$\textsubscript{\SHARP}}

\newcommand{\W}{$\left<w\right>$\xspace}
\newcommand{\A}{$\left<a\right>$\xspace}
\newcommand{\PW}{$\left<p\right>$\xspace}

\newcommand{\literal}[1]{\small{\emph{#1}}}

\usepackage{biblatex}
\title{\THETITLE}

\author{%
  Morris Alper and Hadar Averbuch-Elor
    \\
  Tel Aviv University\\
}

\begin{document}

\maketitle

\begin{abstract}
  Although the mapping between sound and meaning in human language is assumed to be largely arbitrary, research in cognitive science has shown that there are non-trivial correlations between particular sounds and meanings across languages and demographic groups, a phenomenon known as \emph{sound symbolism}. Among the many dimensions of meaning, sound symbolism is particularly salient and well-demonstrated with regards to cross-modal associations between language and the visual domain. In this work, we address the question of whether sound symbolism is reflected in vision-and-language models such as CLIP and Stable Diffusion. Using zero-shot knowledge probing to investigate the inherent knowledge of these models, we find strong evidence that they do show this pattern, paralleling the well-known \emph{kiki--bouba effect} in psycholinguistics. Our work provides a novel method for demonstrating sound symbolism and understanding its nature using computational tools. Our code will be made publicly available\footnote{via our project page \url{https://kiki-bouba.github.io/}}.
\end{abstract}
\section{Introduction}\label{sec:intro}

\emph{``What's in a name? That which we call a rose
by any other name would smell as sweet."}

\qquad  \emph{---William Shakespeare, Romeo and Juliet}\\

Philosophers of language have long debated whether the mapping between sound and meaning in speech is arbitrary. Discussions on this topic date back to the Socratic dialogues of Plato, as well as early modern philosophers such as John Locke and Gottfried Wilhelm Leibniz~\cite{magnus2013history}. Ferdinand de Saussure, the seminal early 20th century linguist and semiotician, famously stated that \emph{le signe est arbitraire}\footnote{``the sign is arbitrary''}. In Saussure's view, words are simply arbitrary conventions and their sounds have no inherent connection to their meanings; hence the French word \emph{chien} and its English translation \emph{dog} both equally denote the same animal despite sharing no sounds in common~\cite{de1916course}.

Although the concept of the arbitrariness of the sign was influential on modern linguistics, there are many evident cases where it does not hold which have attracted great interest among modern researchers. Cases of \emph{sound symbolism} in language, where the sounds themselves in a word have some connection to what they describe, include onomatopoeic phrases such as \emph{kapow} (a punch or banging sound) and \emph{glub-glub} (water bubbling). Beyond direct imitation, it has been noted that English and other languages show correlations between certain phonetic structures and types of meaning, such as the array of English words beginning with \emph{cr-} denoting brittleness (\emph{crush}, \emph{crunch}, \emph{crash}, \emph{crack}, \emph{crackle}, \emph{crinkle}, ...). This raises natural questions including how universal these patterns of iconicity are between languages and cultures, whether they are rooted in psychological tendencies that influence language or vice versa, what acoustic properties of speech sounds influence iconicity, and the extent to which less obvious iconic patterns shape language in general.

\begin{figure}
\centering %
    \centering
    \jsubfig{\includegraphics[height=1.7cm]{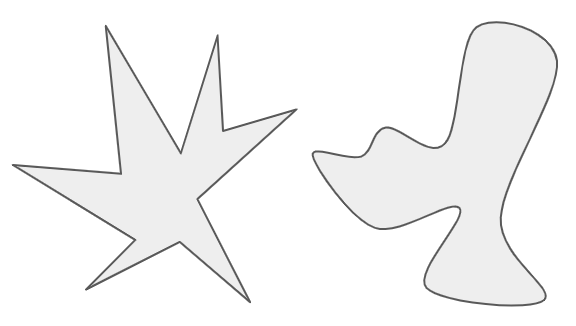}}{Original Experiment}
    \hfill
    \jsubfig{\includegraphics[height=1.7cm]{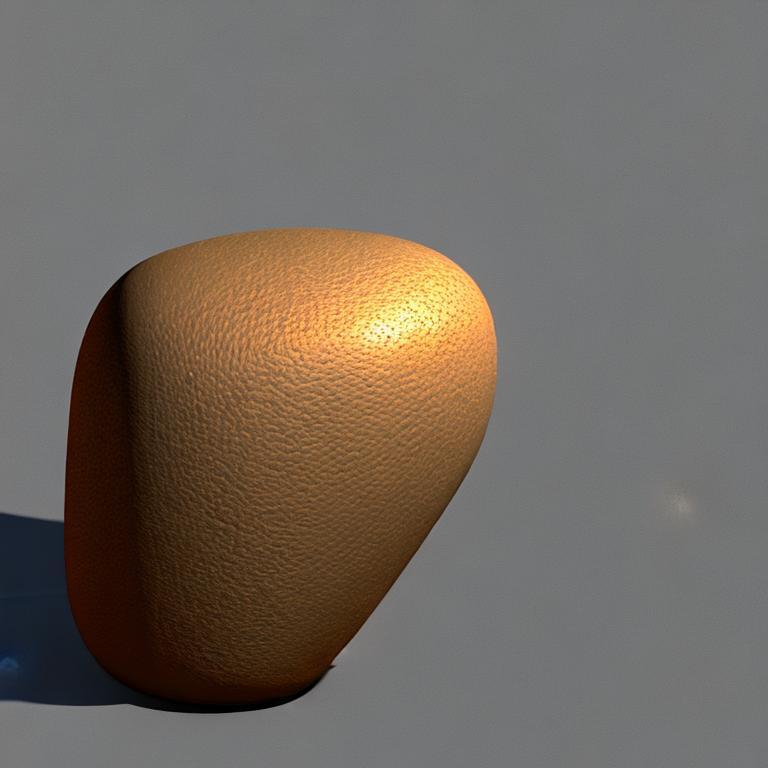}
    \includegraphics[height=1.7cm]{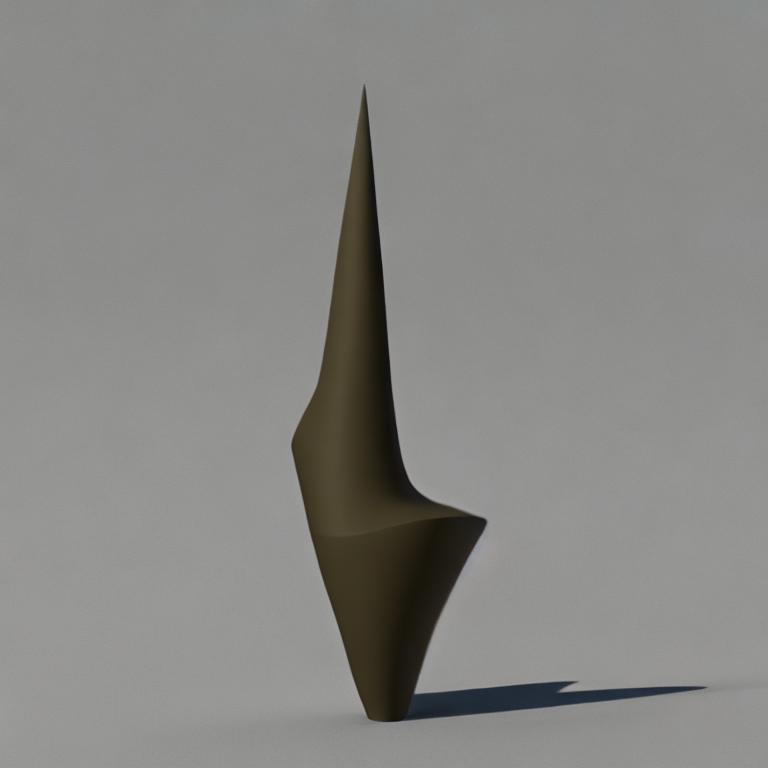}
    \includegraphics[height=1.7cm]{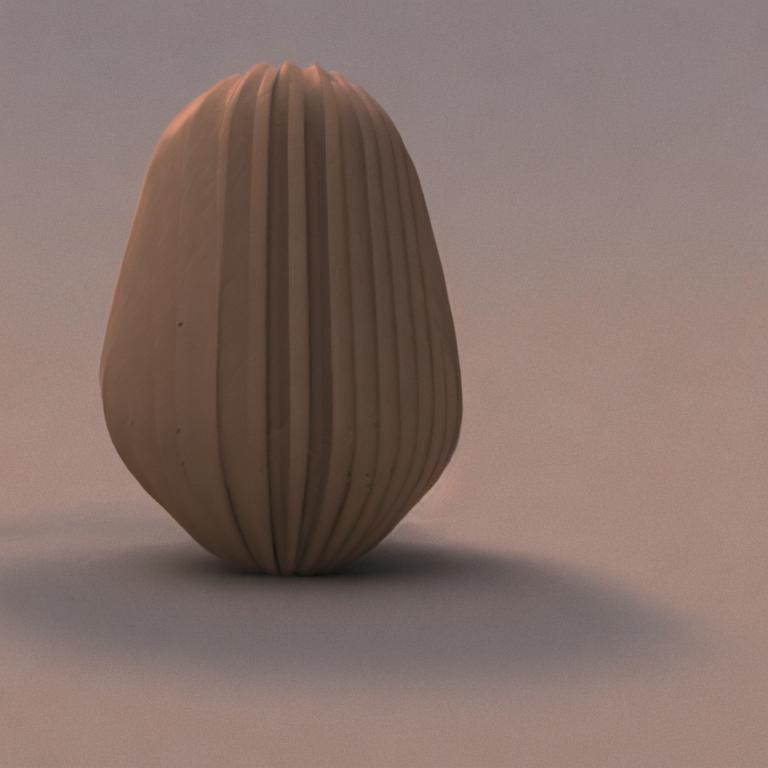}
    \includegraphics[height=1.7cm]{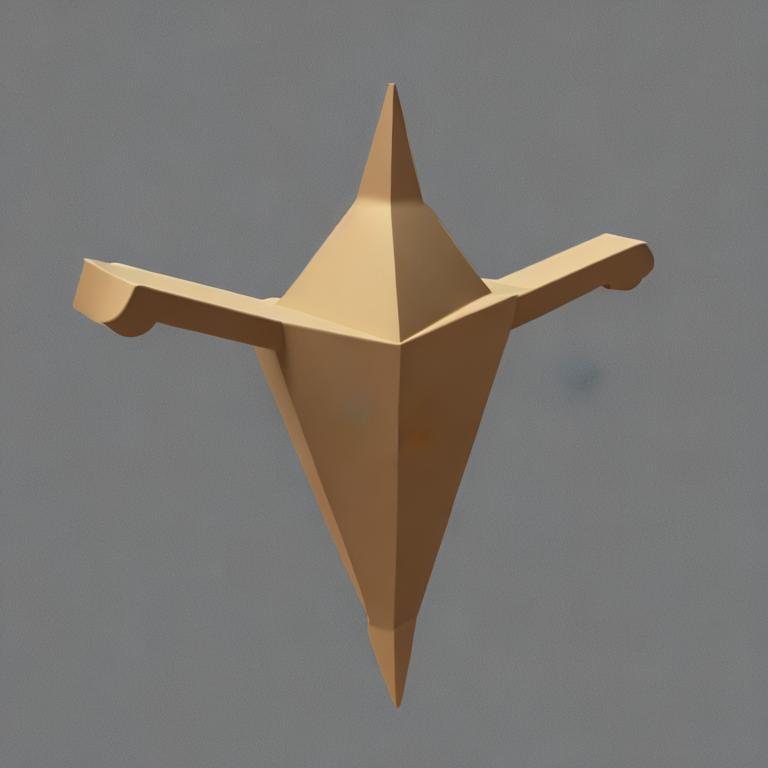}
    \includegraphics[height=1.7cm]{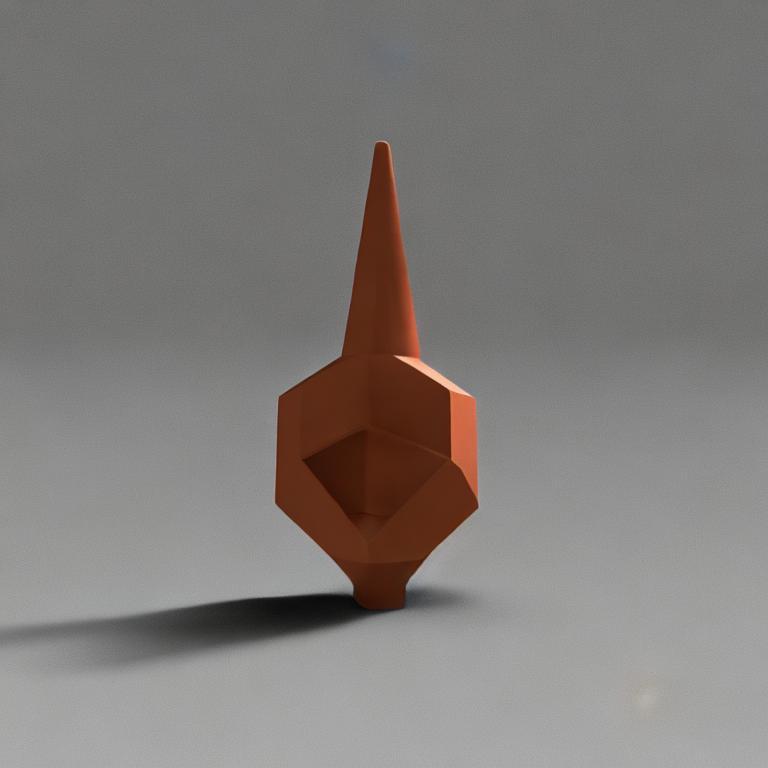}
    \includegraphics[height=1.7cm]{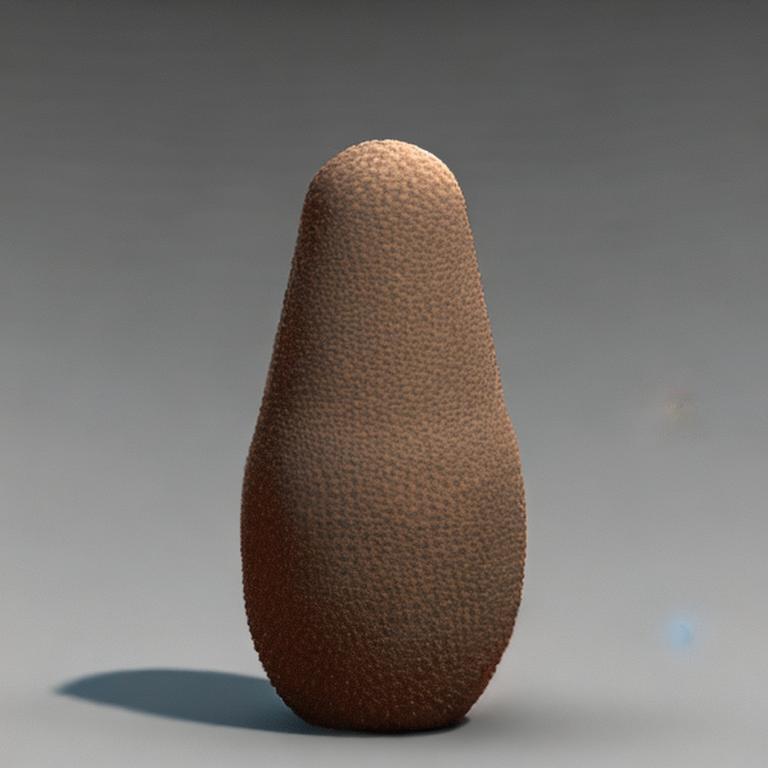}}{Random Image Generations}
    \caption{\textbf{Illustration of the kiki--bouba effect.} The shapes on the far left illustrate stimuli used in the classic \emph{kiki--bouba} experiment. The remaining images are random generations from Stable Diffusion with the prompt \emph{a 3D rendering of a \W shaped object}, where \W $\in$ \{\emph{kiki}, \emph{bouba}\}. Which of of these images do you think were generated using pseudoword \emph{kiki} and which with \emph{bouba}? See below\setcounter{footnote}{2}\protect\footnotemark{} for the answer.}
    \label{fig:teaser}
\end{figure}

\footnotetext{From left to right: \emph{bouba, kiki, bouba, kiki, kiki, bouba}}

Perhaps the most well-known and conclusively demonstrated examples of sound symbolism is the \emph{kiki--bouba effect}. In this experiment, subjects are shown a spiky object and a round object, such as the left-hand shapes in Figure \ref{fig:teaser}. When asked to assign one the name \emph{kiki} and the other the name \emph{bouba}, subjects show an overwhelming preference (see footnote\footnote{The vast majority of subjects prefer the name \emph{kiki} for the spiky object and \emph{bouba} for the round object.} for the preferred matching). This effect has been shown for both auditory stimuli~\cite{fort_resolving_2022} and for written text (which implicitly maps to speech sounds)~\cite{cuskley_phonological_2017,carolis_assessing_2018,cwiek_boubakiki_2022}. Furthermore, an array of studies have demonstrated it to hold across various speech sounds and constructed pseudowords (beyond \emph{kiki} and \emph{bouba})~\cite{mccormick_sound_2015}, between different languages and cultures~\cite{blasi_sound-meaning_2016}, and even among prelingual infants~\cite{ozturk2013sound}. These findings challenge the Saussurean assumption that sound maps purely arbitrarily to meaning.

In another domain, recent years have seen explosive progress in the field of machine learning applied to natural language and vision, mainly powered by transformer neural networks and training on web-scale datasets of captioned images~\cite{bommasani2021opportunities}. This includes discriminative models such as CLIP~\cite{radford2021learning} and generative text-to-image models such as Stable Diffusion~\cite{rombach2022high}. Although these models have revolutionized many applied tasks, they are largely used as a black box. One line of research examines the emergent behavior and inner structure of neural networks in order to interpret their behavior~\cite{rogers2020primer}, and several works investigate the visual knowledge encoded by vision-and-language models (VLMs) in the context of human cognition and visual semantics in language~\cite{alper2023:is-bert-blind,zhang2022visual}. Given these parallels along with fundamental differences between these models and the human brain, it is natural to ask whether they have also learned to associate sounds in language (as reflected in their written representations) with visual semantics. In other words, does the vision-and-language setting reflect the presence of sound symbolic patterns in language? If so, this would provide strong evidence against Saussure's legendary claim from a novel computational perspective.

As seen in Figure \ref{fig:teaser}, generative text-to-image models can be applied to text containing pseudowords such as \emph{kiki} or \emph{bouba}. Consider the various images generated by Stable Diffusion shown on the right-hand side of the figure. The round objects were generated with one of these pseudowords and the sharp objects were generated with the other (see below for the correct mapping), suggesting that the model associates each pseudoword with visual semantic properties such as roundness and sharpness. It is not obvious a priori that VLMs would learn associations with text input that is seemingly not well-formed; this suggestive parallel to human cognition invites methodical study. In this work, we analyze this phenomenon rigorously with quantitative as well as qualitative criteria, including generalizing to a larger set of constructed pseudowords reflecting many different speech sounds in English, and examining correlations between form and meaning on the grapheme (letter) as well as whole-word level.

To test for sound symbolism in VLMs, we use zero-shot knowledge probing which allows for evaluating the models' inherent knowledge---\emph{i.e.,} not new knowledge acquired during further training. We leverage the ability of CLIP to embed text and image data in a shared semantic space in order to probe discriminative and generative (text-to-image) models with the same evaluation metrics. Our tests evaluate whether these models encode pseudowords similarly to humans with respect to known symbolic associations, comparing them to adjectives indicating properties related to ``sharpness'' and ``roundness''. To further ground our results in human cognition, we also conduct a user study testing the ability of subjects to reconstruct pseudowords used to condition text-to-image generations. 
Our results demonstrate that sound symbolism can indeed be observed in VLMs; the models under consideration associate visual meanings with the written forms of particular speech sounds. Although we do not claim to answer exactly how this is learned by these models, our results suggest that VLMs could provide valuable insights regarding sound symbolism. In addition to providing a new perspective on this phenomenon in cognitive science, our work sheds light on what is learned by multimodal models in the context of interpretability in machine learning, demonstrating that these models also encode knowledge over individual characters beyond the semantics of words.

\section{Related Work} \label{sec:rw}

\noindent
\textbf{Sound symbolism in language.} A wide array of studies in cognitive psychology and linguistics have demonstrated the existence of sound symbolic associations, including the famous \emph{kiki--bouba} experiment as well as many variations~\cite{sapir_study_1929,mccormick_sound_2015,lacey_stimulus_2020,adelman_emotional_2018,bottini_sound_2019,drijvers_sound-symbolism_2015,carolis_assessing_2018,cuskley_phonological_2017,dingemanse_what_2016,fort_resolving_2022}. This effect has been shown to be robust cross-culturally and across various languages and writing systems~\cite{blasi_soundmeaning_2016,blasi_sound-meaning_2016,cwiek_boubakiki_2022,bremner_bouba_2013}, and even present among infants and toddlers~\cite{maurer2006shape,ozturk2013sound}. Corpus analyses of English have suggested that sound symbolic trends are present in the basic lexicon of the language -- words with particular meanings may be statistically more likely to contain certain sounds~\cite{monaghan2014arbitrary,winter2021size,sidhu2021sound}. Interestingly, research on blind individuals has found these associations to be weaker or absent when examined between spoken and tactile modalities, suggesting that visual input plays a key role~\cite{fryer2014touching,hamilton2018role}.

\noindent
\textbf{Vision-and-language models.} Recent years have seen rapid development of foundation models, powerful pretrained models that can be adapted to downstream tasks, largely powered by transformer architectures and pretraining on web-scale data~\cite{bommasani2021opportunities}. In the field of multimodal learning, significant foundation models include dual encoder models such as CLIP~\cite{radford2021learning} and its open-source implementation  OpenCLIP~\cite{cherti2022reproducible}, trained with a contrastive objective on captioned image data. The power of these discriminative models for paired text-image understanding has been further leveraged in recent meteoric development of text-to-image generative models. Large diffusion models such as DALL-E 2~\cite{ramesh2022hierarchical} and Imagen~\cite{saharia2022photorealistic} set a new standard for general text conditioned image generation. Latent diffusion models such as Stable Diffusion make comparable results attainable with a smaller computational footprint by applying denoising diffusion to the latent space of an image autoencoder~\cite{rombach2022high}. In particular, the classifier-free guidance~\cite{ho2022classifier} of Stable Diffusion uses OpenCLIP. In our work, we use OpenCLIP and Stable Diffusion as reference SOTA models for discriminative and generative text-image understanding respectively.

\noindent
\textbf{Parallels between VLMs and cognitive science.} A number of prior works have examined the types of knowledge learned by artificial neural networks, with connections to learning and knowledge in cognitive science sometimes left implicit and sometimes stated explicitly. In particular, models trained on vision-and-language tasks have been found to encode knowledge about the visual world in ways that have parallels to human cognition. Although unimodal (text-only) language models learn some degree of factual information~\cite{petroni2019language,rogers2020primer,cohen2023crawling} and commonsense reasoning~\cite{west2021symbolic}, the inclusion of visual data in training may bolster models with visual commonsense knowledge which is typically not written explicitly in text~\cite{vedantam2015learning,lin2015don,kottur2016visual,kojima2020learned,zhang2022visual,alper2023:is-bert-blind}. Alper \emph{et al.}~\cite{alper2023:is-bert-blind} explicitly connect this to similar findings in cognitive science, where studies have explored the effect of human vision impairment and blindness on color associations~\cite{van2021blind, saysani2021seeing, saysani2018colour, shepard1992representation, marmor1978age}. In addition, Orgad \emph{et al.}~\cite{orgad2023editing} note that text-to-image models in particular display implicit assumptions about the world based on correlations and biases in captioned image datasets.

Our work also explores the knowledge learned by VLMs, but rather than investigating the semantics of words or utterances in text, we focus on the \emph{surface form} of textual input to these models and show that they have learnt a non-trivial mapping between sounds encoded by the written text and visual semantics. We also note that the surface form of textual input to these models has been explored in the context of typographic attacks~\cite{goh2021multimodal,materzynska2022disentangling}, in which models learn to associate text with images containing the text written out visually; however, they do not address semantic associations with the surface form of the input beyond its literal appearance as text.
\section{Computational Paradigm for Sound Symbolic Probing} \label{sec:exp}

To test for the presence of sound symbolism in VLMs, we design a paradigm using controlled comparisons between pseudowords -- nonsense words constructed from English letters with desired properties -- and visual semantic properties. In particular, we are interested in whether a VLM associates pseudowords with known sound-symbolic associations among humans (e.g. \emph{kiki} -- sharp) with visual properties such as ``sharpness''. For generative (text-to-image) models this translates to testing whether pseudowords containing letters that are known to have ``sharp'' associations tend to generate sharper images on average, and conversely for pseudowords containing letters with ``round'' associations. Recent progress in the field of multimodal learning makes it possible to test this directly, as CLIP~\cite{radford2021learning} can be used to test images for semantic properties by comparing them to text prompts (e.g. \emph{a 3D rendering of a sharp object}) via cosine similarity of embedding vectors in a semantic space shared between the textual and visual modalities. Moreover, CLIP can be tested directly as a discriminative model for sound symbolism by comparing the text embeddings of pseudowords to those of such properties. 

This intuition is formalized below. We first describe how pseudowords with ground truth associations are constructed (Section \ref{sec:pseudo}). We then elaborate on our zero-shot knowledge probing techniques (Section \ref{sec:probe}) and evaluation protocol (Section \ref{sec:eval-protocol}). %

\subsection{Pseudoword Construction} \label{sec:pseudo}

Although studies using particular pseudoword stimuli such as \emph{kiki--bouba}~\cite[p.19]{ramachandran2001synaesthesia} or \emph{maluma--takete}~\cite[p.133]{kohler1947gestalt} have found that certain speech sounds show cross-modal associations, the precise nature of which acoustic phenomena give rise to these associations is an active topic of research. McCormick \emph{et al.}~\cite{mccormick_sound_2015} demonstrate that speech sounds form a spectrum from most ``sharp'' to most ``round''. Relevant phonetic distinctions include sonority and voicing in consonants, and vowel height and rounding. Fort and Schwartz~\cite{fort_resolving_2022} suggest that these properties can be aligned on the dimensions of ``spectral balance and temporal continuity'', which have also been found relevant for sharpness associations with non-speech sounds such as drum beats~\cite{fort_resolving_2022} and electronic sound waves~\cite{parise2009birds}.

To test for the presence of sound symbolism in the VLMs under consideration, we split speech sounds into ``highly sharp'' and ``highly round'' categories based on phonetic properties, leaving an investigation of more fine-grained distinctions within the spectrum of sounds to future research. Following the classification of speech sounds described in \cite{mccormick_sound_2015}, we define the following subsets of English graphemes (letters) for use in our experiments, divided between consonants and vowels and split into classes based on their corresponding sounds:

\begin{tabular}{llllllll}
     \CS: & \graphemic{p t k s h x} & & \CR: & \graphemic{b d g m n l} & & &  \\
     \VS: & \graphemic{e i} & & \VR: & \graphemic{o u} & & \VN: & \graphemic{a}
\end{tabular}

The classes above use \SHARP and \ROUND as shorthand for ``sharp'' and ``round'' associations respectively, and \VN indicates a neutral association (may appear in either class of pseudoword, as described below). The main phonetic distinctions used to determine these classes are voicing for consonants and backness for vowels; see the supplementary material for more information about the phonetic details motivating these classes.

We construct pseudowords for use in our experiments using the three syllable template (CV)\textsubscript{1}(CV)\textsubscript{2}(CV)\textsubscript{1} , where the first and last syllables are the same (e.g. \emph{kitaki}, \emph{bodubo}, \ldots). In addition, all graphemes must be drawn either from \CSnospace$\cup$\VSnospace$\cup$\VN or \CRnospace$\cup$\VRnospace$\cup$\VN, disallowing forms like \emph{kiduki} which mix graphemes from the two classes. We denote the set of all pseudowords with graphemes from the former set as \PseudoS, the set of pseudowords with graphemes from the latter set as \PseudoR, and \Pseudo$\,=\,$\PseudoSnospace$\,\cup\,\,$\PseudoR.
The choice of this template has a few motivations, including the large number of possible forms, few clashes with existing English words, and similarity to the \emph{maluma--takete} stimuli used by Köhler~\cite[p.133]{kohler1947gestalt}. Examples of such pseudowords include the following (in arbitrary order):

\begin{tabular}{llllllllll}
     \PseudoS: & \emph{kitaki} & \emph{hatiha} & \emph{pepape} & \emph{xisixi} & \emph{hipehi} & \emph{xaxaxa} & \emph{texete} & \emph{...} & (324 total)\\
     \PseudoR: & \emph{gugagu} & \emph{bodubo} & \emph{gunogu} & \emph{daluda} & \emph{momomo} & \emph{lunulu} & \emph{gadaga} & \emph{...} & (324 total) %
\end{tabular}

We also report results for the pseudowords \emph{kiki} and \emph{bouba} used verbatim, to reproduce the most well-known form of the \emph{kiki--bouba} effect with VLMs. In accordance with the principles used to construct our pseudowords, \emph{kiki} corresponds to class \SHARP and \emph{bouba} corresponds to class \ROUND.

\subsection{Zero-shot Knowledge Probing} \label{sec:probe}

\raggedbottom

We probe VLMs for sound symbolism with a zero-shot linear probe applied to embedding vectors in the multimodal embedding space of CLIP ($\subset \mathbb{R}^{1024}$). We consider only the zero-shot regime since we are interested in the inherent knowledge of our models (acquired from pretraining) and not the dynamics of training on new data. Furthermore, our approach tests VLMs end-to-end and is agnostic to the source
of these effects with respect to the relative contribution of different model-internal components (such as tokenization and hidden activations).

\noindent
\textbf{Prompts used}: We use the following prompts to probe the models under consideration, where \W is the item (word or pseudoword) to be inserted into the prompt:

\begin{tabular}{ll}
     $P_1$: & ``\emph{a 3D rendering of a \W object}'' \\
     $P_2$: & ``\emph{a 3D rendering of a \W shaped object}'' 
\end{tabular}

We use $P_1$ for adjectives (e.g. \emph{...of a round object}) and $P_2$ for nouns and pseudowords (e.g. \emph{...of a cactus shaped object}, \emph{...of a kiki shaped object}). These prompts are chosen for visual simplicity and interpretability; in the supplementary material we compare results on other prompts.

\noindent
\textbf{Embeddings}: All of our tests use embedding vectors in CLIP space ($\subset \mathbb{R}^{1024}$) corresponding to pseudowords and adjectives inserted into prompts. These may either be text embeddings calculated using CLIP's text encoder, or image embeddings using CLIP's vision encoder applied to image generations conditioned on the given text. In either case, we embed pseudowords and adjectives in CLIP space to obtain vectors $v_{\text{\W}}, w_{\text{\A}} \in \mathbb{R}^{1024}$ respectively and unit-normalize them to $\hat{v}_{\text{\W}}, \hat{w}_{\text{\A}}$. When evaluating CLIP we directly encode both as text; when evaluating Stable Diffusion we calculate $v_{\text{\W}}$ using image generation, as detailed further in Section \ref{sec:models}.

\noindent
\textbf{Geometric scores} $\gamma_{\text{\W}}$: We propose a scoring method to measure geometric attributes such as sharpness or roundness, applied to images of objects or embeddings corresponding to pseudowords. To calculate this, we identify the one-dimensional semantic direction of interest in CLIP embedding space aligned with a collection of adjectives, similar to prior work on finding interpretable linear subspaces of word embeddings~\cite{mikolov2013efficient,bolukbasi2016man,dev2019attenuating} and multimodal embedding spaces~\cite{tewel2021zero, patashnik2021styleclip}. We manually select 20 ground-truth adjectives, split evenly between those with sharp or round associations which we denote by \AdjS and \AdjR respectively. These are as follows:

\begin{tabular}{lll}
     \AdjS & $=$ & $\{$\emph{sharp, spiky, angular, jagged, hard, edgy, pointed, prickly, rugged, uneven}$\}$ \\
     \AdjR & $=$ & $\{$\emph{round, circular, soft, fat, chubby, curved, smooth, plush, plump, rotund}$\}$
\end{tabular}

Using the embeddings of these adjectives, we construct probe vector $w_{adj} := \sum_{\text{\A} \in \text{\AdjS}} \hat{w}_{\text{\A}} - \sum_{\text{\A} \in \text{\AdjR}} \hat{w}_{\text{\A}}$, unit-normalized to $\hat{w}_{adj}$. This vector approximates the direction in CLIP space representing the visual semantic dimension of interest; in other words, this is a unit vector pointing in the direction in CLIP’s latent space which distinguishes between the two sets of adjectives. We probe item \W by calculating the score $\gamma_{\text{\W}} := \hat{v}_{\text{\W}} \cdot \hat{w}_{adj}$, corresponding to projection onto the 1D subspace spanned by $w_{adj}$. Intuitively, the scalar $\gamma_{\text{\W}}$ measures whether the given item is closer to the ``round'' or ``sharp'' end of the scale of associations in our model's semantic space. Tangentially, while this work focuses on using these scores to place pseudowords along a sharp--round semantic axis, our geometric scoring method could be applicable for identifying words or images with such associations in more general settings as well.
\noindent

\textbf{Phonetic scores} $\phi_{\text{\W}}$: We also propose a complementary scoring method to measure associations with categories of sounds reflected in English spelling. This is applied in order to quantify the phonetic or graphemic associations that our models show with real English words such as the adjectives given above. We construct probe vector $v_{pw} := \sum_{\text{\W} \in \text{\PseudoS}} \hat{v}_{\text{\W}} - \sum_{\text{\W} \in \text{\PseudoR}} \hat{v}_{\text{\W}}$, unit-normalized to $\hat{v}_{pw}$. This is a unit vector which estimates the dimension in CLIP space that best distinguishes between the two ground-truth classes of pseudowords. We then score item \W via cosine similarity with this probe: $\phi_{\text{\W}} := \hat{v}_{pw} \cdot \hat{w}_{\text{\W}}$. Intuitively, the scalar $\phi_{\text{\W}}$ measures whether the given item (\emph{e.g.}, adjective) is closer to pseudowords like \emph{kitaki}\textsubscript{\SHARP} or to those like \emph{bodubo}\textsubscript{\ROUND} in our model's semantic space. We call this ``phonetic scoring'' because it is a semantic dimension determined solely by the letters used in pseudowords and their underlying sounds.

\subsection{Evaluation Method} \label{sec:eval-protocol}

For quantitative analysis, we evaluate the scores $\gamma_{\text{\W}}$ and $\phi_{\text{\W}}$ using classification metrics; $\gamma_{\text{\W}}$ for predicting the binary class (\SHARP or \ROUND) of pseudowords, and $\phi_{\text{\W}}$ for predicting the binary class (\SHARP or \ROUND) of adjectives with ground-truth labels. As these are unnormalized scores, we use non-probabilistic and threshold-agnostic metrics, namely ROC-AUC and Kendall correlation $\tau$ between the scores in question and the ground-truth classes as a binary indicator variable. These also have the desirable property of being symmetric with respect to our class labels, neither of which is privileged with respect to the other. Additionally, we provide an analysis of the correlation between $\gamma_{\text{\W}}$ and the particular sounds present in the first syllable of the given pseudoword.

We also reproduce the classic \emph{kiki--bouba} experiment using the scores $\gamma_{kiki}$ and $\gamma_{bouba}$. To place these scores on an interpretable scale, we use their percentile ranks relative to scores of all pseudowords in \Pseudo, and report the difference in these percentiles $\Delta P_{kb}$. A high value for this metric indicates that the given model strongly aligns these pseudowords along the semantic axis of roundness and sharpness.

\section{Results and Evaluation}

In this section, we present our main findings. We first discuss details of the experimental setup (Section \ref{sec:models}). We then present quantitative results (Section \ref{sec:quant}), the results of our user study (Section \ref{sec:user_study}), and qualitative results (Section \ref{sec:qual}). %

Further experimental details and results are provided in the supplementary material, including results on additional prompts and model architectures (such as a GAN-based text-to-image model). All of these settings show results consistent with our primary findings on CLIP and Stable Diffusion. There we also provide results for unimodal (text-only) text encoder models, which show mixed results when probed for sound symbolism.

Our primary findings in this work use models trained and prompted with texts in the English language. Nonetheless, in the supplementary material we also probe for sound symbolism in a multilingual text-to-image model with prompts in four geographically and linguistically diverse languages, with positive results. While we do not directly address the universality of sound symbolism across languages, these results warrant further investigation of sound symbolism in multilingual VLMs.

\subsection{Experimental Details} \label{sec:models}

We investigate the presence of sound symbolic patterns in two types of VLM: generative text-to-image models as exemplified by Stable Diffusion~\cite{rombach2022high}, and discriminative models with dual text and vision encoders as exemplified by CLIP~\cite{radford2021learning}. We use the open-source CLIP implementation OpenCLIP~\cite{cherti2022reproducible} throughout; Stable Diffusion uses OpenCLIP as its text encoder for classifier-free guidance, motivating this comparison. We use these models as-is and probe them in the zero-shot regime, without any further training or adjustment of their tokenizer (which is identical for both models).

We note the a priori possibility that Stable Diffusion's weights may encode additional visual commonsense that is not present in CLIP. The majority of its parameters are in its UNet component (866M vs. 340M in its CLIP text encoder). These weights are trained on an image denoising task which requires understanding local regions in images, while CLIP's pretraining knowledge is only acquired from global image-text matching.

For CLIP, we assign embeddings to pseudowords by inserting them into textual prompts and embedding them with CLIP's text encoder. For Stable Diffusion, we embed pseudowords by generating images using such prompts and embedding the generated images using CLIP's vision encoder. Since the image generation process is stochastic, we reduce variance by generating multiple images with the corresponding prompt and calculating their mean embedding vector. As CLIP embeds text and images in the same space, we evaluate the pseudoword embeddings yielded by both models with the same probing method, as described in Section \ref{sec:eval-protocol}.

\subsection{Quantitative Evaluation} \label{sec:quant}

\begin{table}[t]
  \centering
  \setlength{\tabcolsep}{5.5pt}
  \def\arraystretch{0.95}
  \begin{tabularx}{0.6\columnwidth}{lcccccc}
    \toprule
    &
    \multicolumn{3}{c}{$\gamma_{\text{\W}}$} & &
    \multicolumn{2}{c}{$\phi_{\text{\W}}$}
    \\
    \cmidrule(lr){2-4}
    \cmidrule(lr){6-7}
    Model & AUC & $\tau$ & $\Delta P_{kb}$ & & AUC & $\tau$  \\
    \midrule
    Stable Diffusion & 0.74 & 0.34 & 80\% & & 0.97 & 0.68 \\
    CLIP & 0.77 & 0.39 & 52\% & & 0.98 & 0.70 \\ %
    \midrule
    (random) & 0.50 & 0.00 & 0\% & & 0.50 & 0.00 \\
    \bottomrule
  \end{tabularx}
  \vspace{5pt}
  \caption{\textbf{Results of zero-shot linear probing.} Results under $\gamma_{\text{\W}}$ indicate metrics for predicting pseudoword class (\SHARP or \ROUND) from geometric scores; results under $\phi_{\text{\W}}$ indicate metrics for predicting adjective class (\SHARP or \ROUND) from phonetic scores. Probing methods and evaluation metrics are as described in Section \ref{sec:exp}. (random) indicates the expected performance of a purely random scoring method, providing a lower bound for the performance of the models under consideration.}
\label{tab:association_metrics}
\end{table}

\begin{figure}
    \centering
    \jsubfig{\includegraphics[height=2.4cm]{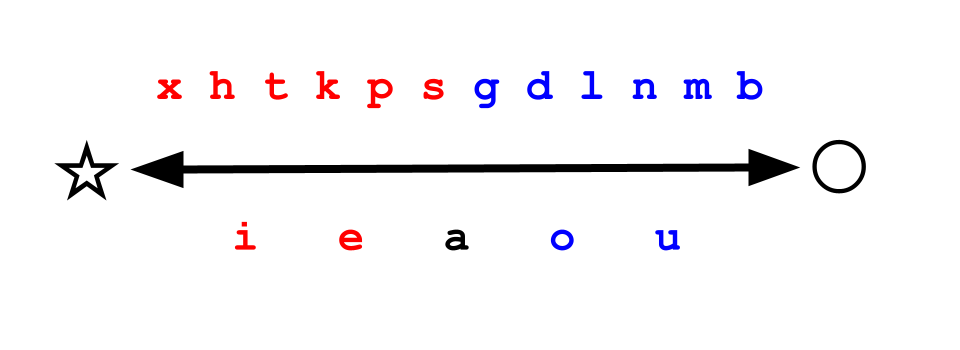}}{\vspace{-15pt}\textbf{Stable Diffusion}}
    \hspace{8pt}
    \jsubfig{\includegraphics[height=2.4cm]{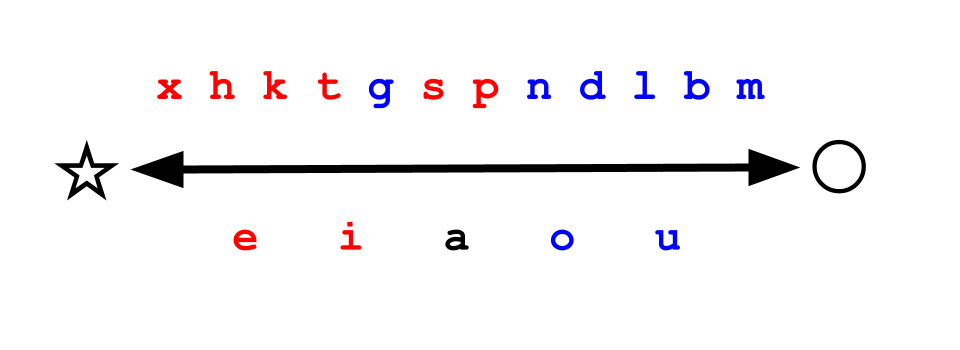}}{\vspace{-15pt}\textbf{CLIP}}
    \caption{\textbf{Graphemes sorted by average geometric score $\gamma_{\text{\W}}$} for pseudowords \W whose first syllable contains the given grapheme, calculated with Stable Diffusion and CLIP. Characters are colored based on their ground-truth association (\textcolor{red}{red} for \SHARP, \textcolor{blue}{blue} for \ROUND). Consonants are shown above and vowels below the arrow. We see that the two classes are mostly well-discriminated by these scores, especially when calculated Stable Diffusion. In this visualization, consonants and vowels are displayed on separate scales and are not positioned absolutely with respect to each other.}
    \label{fig:discrim_letters}
\end{figure}

\begin{figure}
    \centering
    \jsubfig{\includegraphics[height=2.4cm]{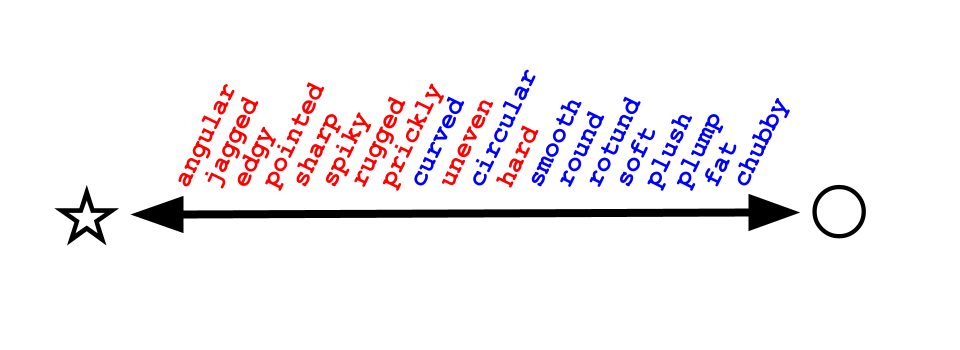}}{\vspace{-15pt}\textbf{Stable Diffusion}}
    \hspace{8pt}
    \jsubfig{\includegraphics[height=2.4cm]{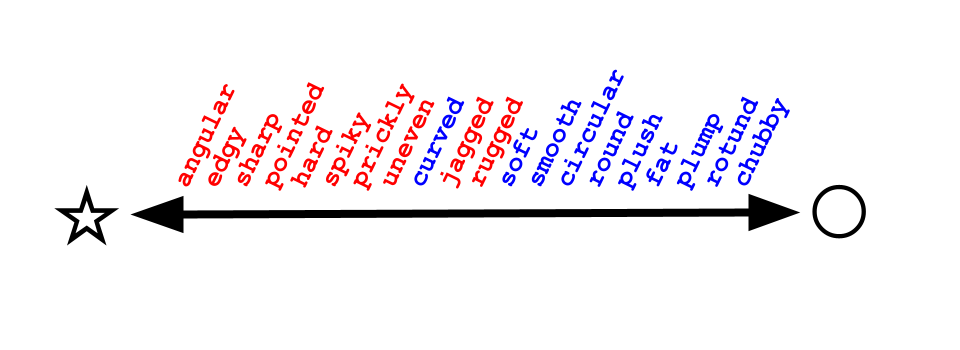}}{\vspace{-15pt}\textbf{CLIP}}
    \caption{\textbf{Ground-truth adjectives sorted by phonetic score $\phi_{\text{\W}}$}, calculated with Stable Diffusion and CLIP. Adjectives are colored based on their ground-truth association (\textcolor{red}{red} for \SHARP, \textcolor{blue}{blue} for \ROUND). We see that the two classes are highly differentiated by phonetic score for both models, as further reflected in the corresponding metrics in Table \ref{tab:association_metrics}.}
    \label{fig:adj_scale}
\end{figure}

We present the results of our quantitative tests in Table \ref{tab:association_metrics}. Across all metrics, our probing methods are able to predict pseudoword and adjective classes (\SHARP or \ROUND) significantly better than chance. In particular, the results on Stable Diffusion indicate that images generated from pseudowords in \PseudoS are more likely to be visually ``sharp'' and images generated from pseudowords in \PseudoR are more likely to be visually ``round'', consistent with the examples shown in Figure \ref{fig:gen_examples}. Additionally, we see that the $\Delta P_{kb}$ scores indicate a replication of the \emph{kiki--bouba} effect in multimodal models, with \emph{kiki} having a relatively higher affinity for ``sharp'' adjectives and the latter for ``round'' adjectives in such models.

In Figure \ref{fig:discrim_letters}, we see graphemes (letters), split into consonants and vowels and sorted by the average geometric score $\gamma_{\text{\W}}$ of all pseudowords containing them in the first syllable (first two characters).
Surprisingly, we see an emergent pattern that closely reflects psychological and phonetic phenomena, as both models perfectly or nearly perfectly differentiate between \SHARP and \ROUND-associated graphemes (as described in Section \ref{sec:pseudo}). This is despite the fact that these models were trained on text-image pairs on the full caption level, primarily saw valid English text during training, and did not have direct access to the auditory modality at all. We even see intriguing patterns within the intra-class ordering of graphemes, such as close vowels \graphemic{i u} having more extreme association than close-mid vowels \graphemic{e o} and voiced sonorants \graphemic{m n l} having among the most ``round'' associations, although we leave investigation of such fine-grained patterns to future research.

Figure \ref{fig:adj_scale} shows the ground-truth adjectives from \AdjS and \AdjR sorted by phonetic score for both models, corresponding to the $\phi_{\text{\W}}$ metrics in Table \ref{tab:association_metrics}. The near-perfect discrimination between the two classes under both models is consistent with the high metric values seen in the table.

\subsection{User Study} \label{sec:user_study}

In order to ground our results in human cognition, we conduct a user study using images produced by Stable Diffusion prompted using pseudowords. We adopt the two-alternative forced choice paradigm, where participants are provided with image pairs and corresponding pairs of pseudowords, and are asked to match the pseudowords to the corresponding images. We perform this study in two settings: in the first setting, we test participants' ability to distinguish between images generated using the specific pseudowords \emph{kiki} and \emph{bouba}; in the second setting, we test their ability to distinguish between images generated using random pseudowords from \PseudoS and \PseudoR. One hundred subjects participated in the first setting, and seventy five participated in the second setting. Participants provided correct answers with overall accuracy of 73\% in the \emph{kiki--bouba} setting and 55\% in the general pseudoword setting. In order to determine the statistical significance of these results while accounting for variation between subjects and items within each survey, we adopt a mixed-effects logistic regression model. We regress whether a question is answered correctly while treating question identity as a fixed effect and respondent identity as a random effect. Analysis of the first setting (\emph{kiki} vs. \emph{bouba}) results in an intercept estimate corresponding to a 89\% overall success probability ($p < 0.001$); in the second setting (random pseudowords) this results in an intercept estimate corresponding to a 78\% overall success probability ($p < 0.001$). By isolating the overall success rate from the effects of individual respondents and questions, our mixed-effects model results indicate that the sound symbolism exhibited by our text-to-image model correlates with human sound symbolic associations (with a stronger effect size in the simpler setting, but having a significant effect in both settings). Please refer to the supplementary material for additional details, including participant sourcing and compensation, the full text of instructions and survey examples, and a full statistical analysis of survey results.

\subsection{Qualitative Results} \label{sec:qual}

\begin{table}[t]
  \centering
  \setlength{\tabcolsep}{5.5pt}
  \def\arraystretch{0.95}
  \begin{tabularx}{0.99\columnwidth}{lll}
    \toprule
    POS & Lowest $\phi_{\text{\W}}$ ($\Rightarrow$ \ROUND) & Highest $\phi_{\text{\W}}$ ($\Rightarrow$ \SHARP)\\
    \midrule
    \multicolumn{3}{l}{\textbf{Stable Diffusion}} \\
     Noun & \literal{butterball, yolk, pregnancy, booger, eggnog,} & \literal{shard, kite, origami, hexagon, diamond, flake,} \\
    & \literal{turnip, bellyful, crybaby, doughboy} & \literal{octagon, triangle, protractor, lozenge, foldout} \\
    \rule{0pt}{3ex}
    Adj. & \literal{obese, chubby, stinky, pudgy, overweight,} & \literal{triangular, diagonal, angular, shattering, jagged,} \\
    & \literal{fat, pregnant, plump, drowsy, soggy, squishy} & \literal{rectangular, edgy, housebroken, geometrical} \\
    \midrule
    \textbf{CLIP} \\
    Noun & \literal{doughboy, loudmouth, gumdrop, boogeyman,} & \literal{prefix, talkativeness, asker, flexibility, shears,} \\
    & \literal{madwoman, lord, butterball, goddaughter} & \literal{shift, peek, slope, task, exit, hemline, tightness} \\ %
    \rule{0pt}{3ex}
    Adj. & \literal{muggy, soggy, gloomy, grouchy, lumpy} & \literal{apelike, flexible, diagonal, static, triangular,} \\
    & \literal{humongous, hungry, cloudless, unsmiling} & \literal{external, shipshape, interlocking, angular} \\ %
    \bottomrule
  \end{tabularx}
  \vspace{5pt}
  \caption{\textbf{Real English words sorted by phonetic score $\phi_\text{\W}$}. Results are split by part of speech (POS), from the ${\sim}5.5K$ nouns in \Noun and ${\sim}1K$ adjectives in \Adj. Columns indicate items with the lowest and highest scores out of the entire sorted lists, corresponding to more relative similarity to the pseudowords in \PseudoR or \PseudoS respectively. As seen above, these clearly depict ``round'' and ``sharp'' characteristics, particularly the words selected with the generative pipeline (\emph{i.e.} Stable Diffusion).}
\label{tab:sorted_real_words}
\end{table}

We provide a qualitative analysis using real words from English, filtered to select for basic words and split by part of speech: ${\sim}5.5K$ nouns denoted by \Noun, and ${\sim}1K$ adjectives denoted by \Adj. We select basic concrete words for both categories by using lemmas from WordNet~\cite{miller1995wordnet,miller1998wordnet}, filtering out obscure or highly abstract items using age of acquisition and concreteness scores from \cite{kuperman2012age,brysbaert2014concreteness}.
We sort these items by their $\phi_{\text{\W}}$ scores and display the words with the highest and lowest scores.

\begin{figure}
    \centering

    \begin{tabular}{c}
        \rotatebox{90}{High $\gamma_{\text{\W}}$} \rotatebox{90}{$\;\;$($\Rightarrow$ \SHARP)}
        \hspace{5pt}
        \jsubfig{\includegraphics[height=1.7cm]{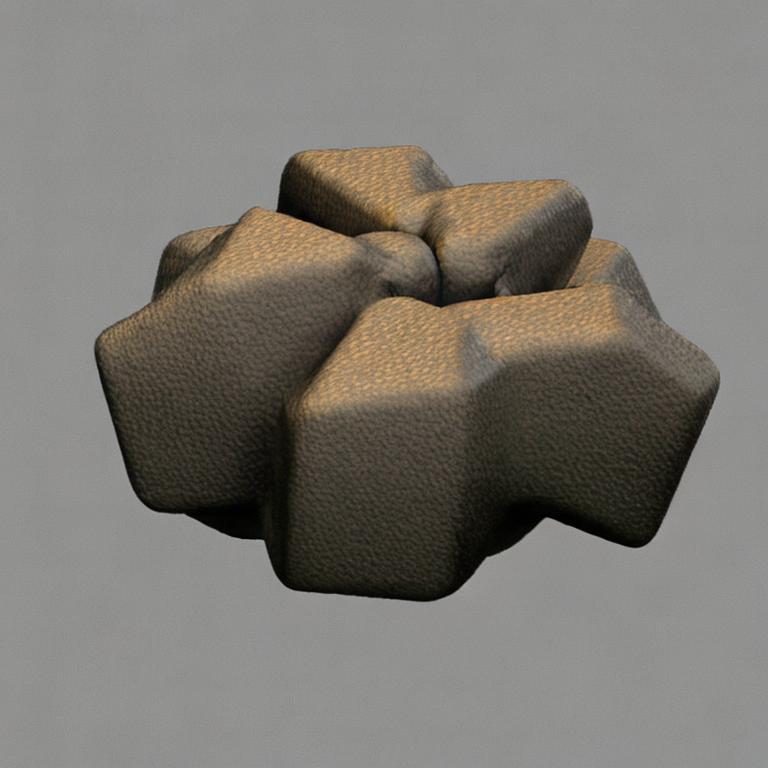}}{\emph{tepite}\textsubscript{\SHARP}}
        \hspace{8pt}
        \jsubfig{\includegraphics[height=1.7cm]{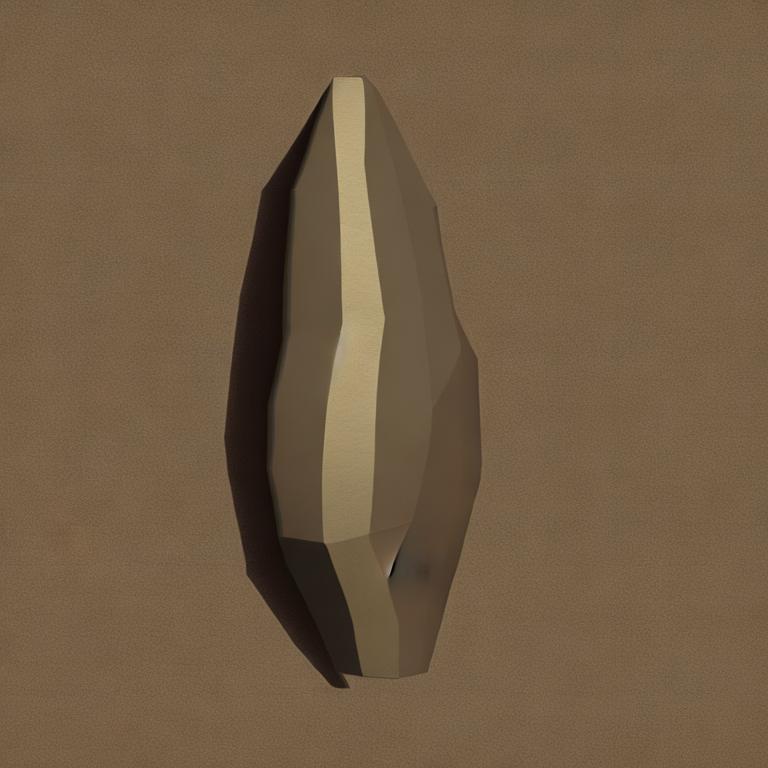}}{\emph{kataka}\textsubscript{\SHARP}}
        \hspace{8pt}
        \jsubfig{\includegraphics[height=1.7cm]{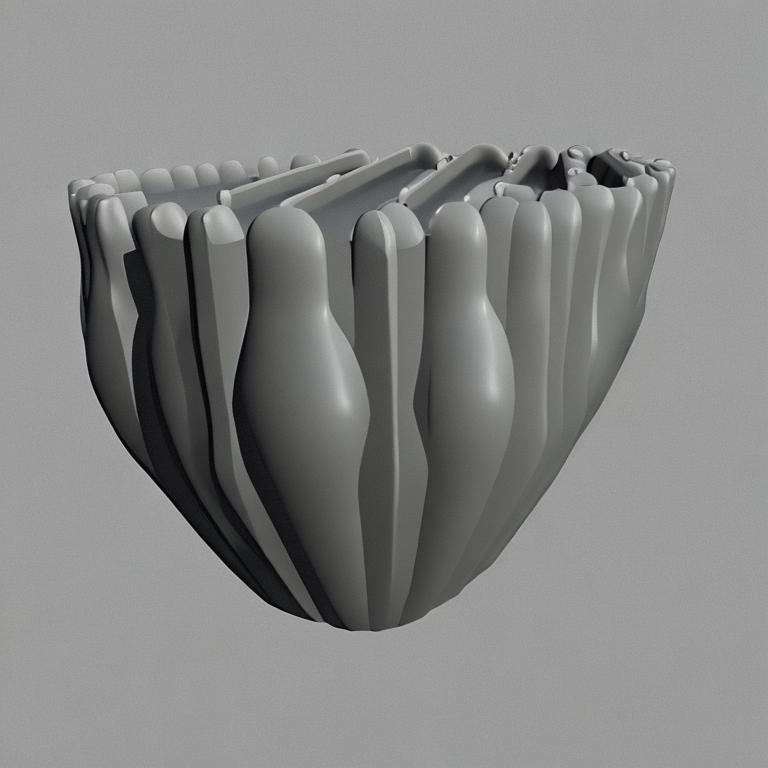}}{\textcolor{red}{\emph{gugugu}\textsubscript{\ROUND}}}
        \hspace{8pt}
        \jsubfig{\includegraphics[height=1.7cm]{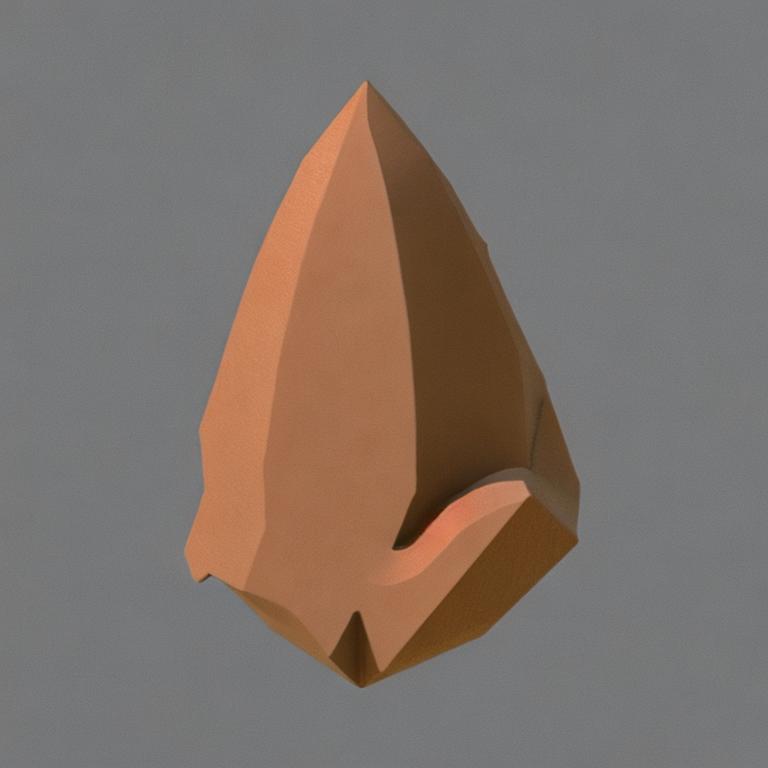}}{\emph{pitapi}\textsubscript{\SHARP}}
        \hspace{8pt}
        \jsubfig{\includegraphics[height=1.7cm]{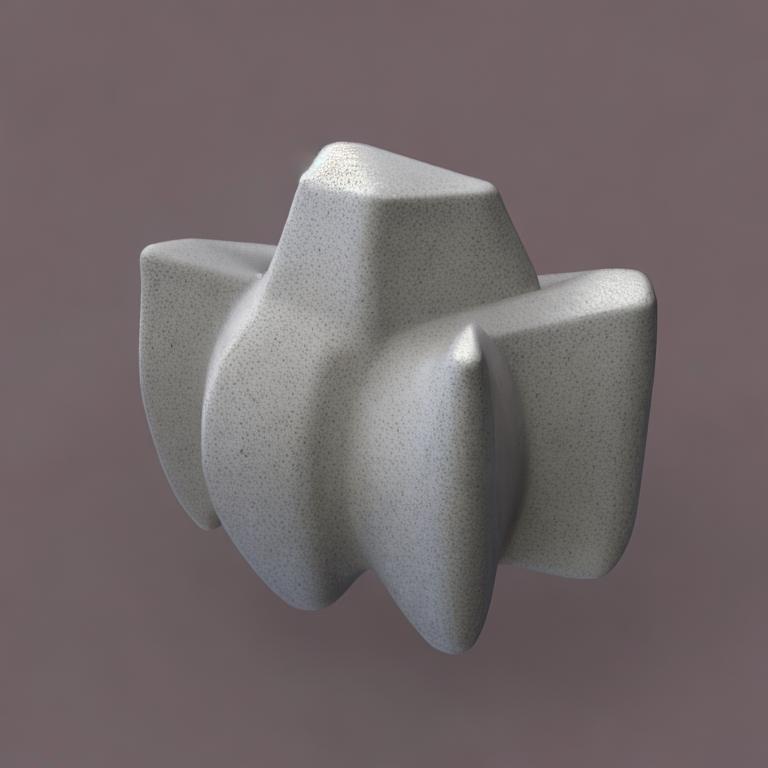}}{\textcolor{red}{\emph{daguda}\textsubscript{\ROUND}}}
        \hspace{8pt}
        \jsubfig{\includegraphics[height=1.7cm]{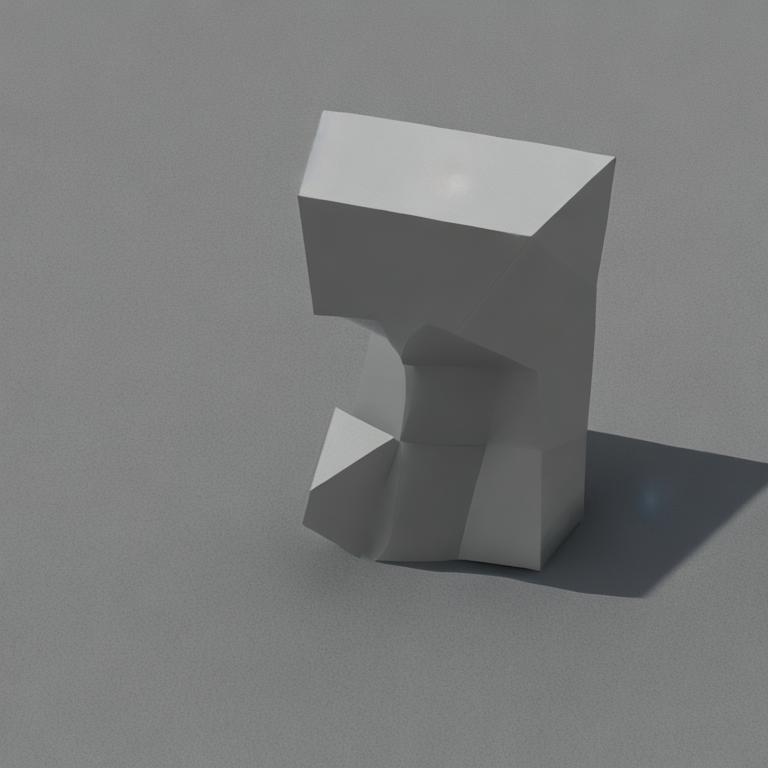}}{\emph{setese}\textsubscript{\SHARP}}
        \\
        \\
        \rotatebox{90}{Low $\gamma_{\text{\W}}$} \rotatebox{90}{$\;\;$($\Rightarrow$ \ROUND)}
        \hspace{5pt}
        \jsubfig{\includegraphics[height=1.7cm]{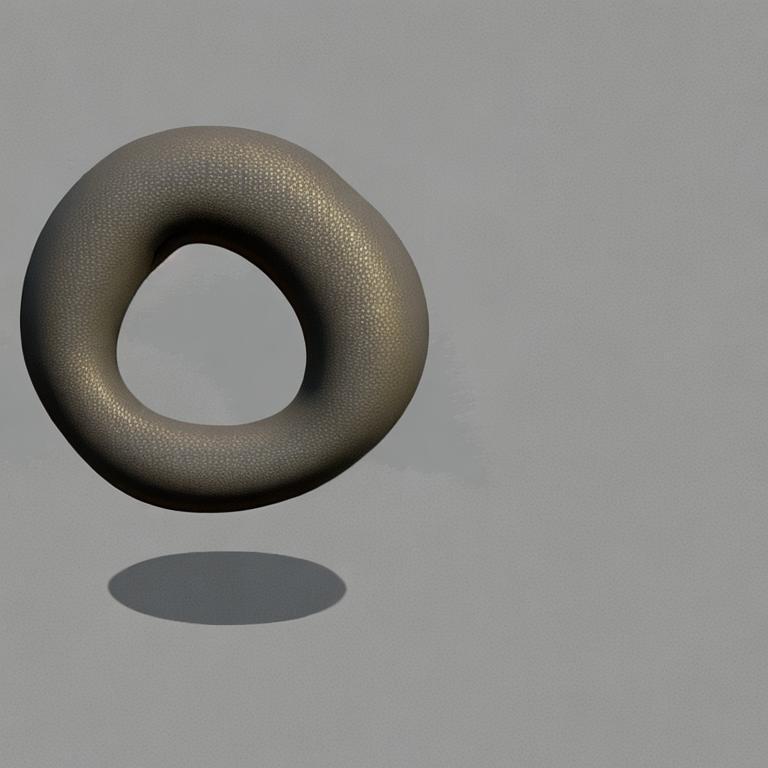}}{\emph{nubanu}\textsubscript{\ROUND}}
        \hspace{8pt}
        \jsubfig{\includegraphics[height=1.7cm]{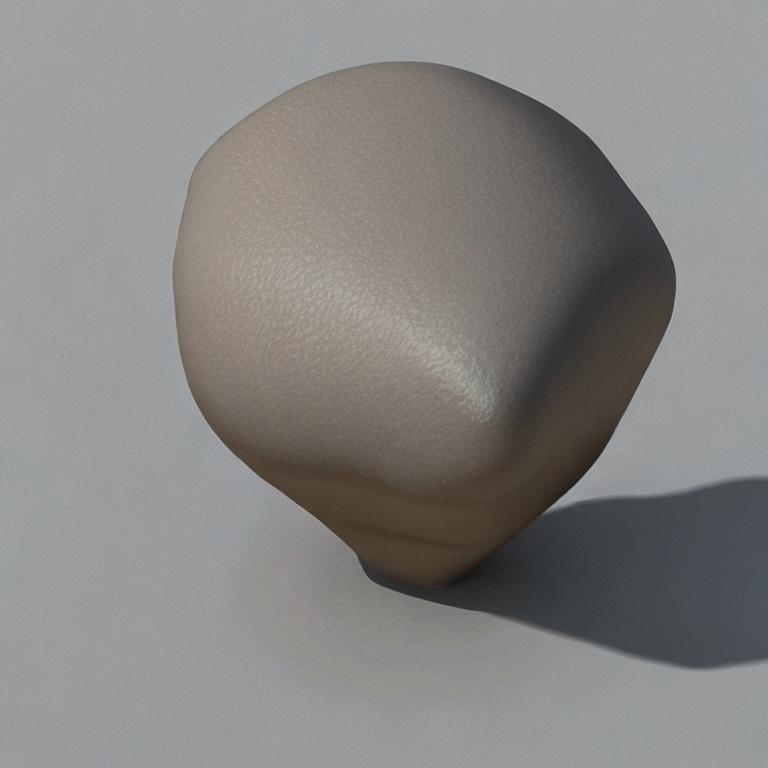}}{\emph{gomago}\textsubscript{\ROUND}}
        \hspace{8pt}
        \jsubfig{\includegraphics[height=1.7cm]{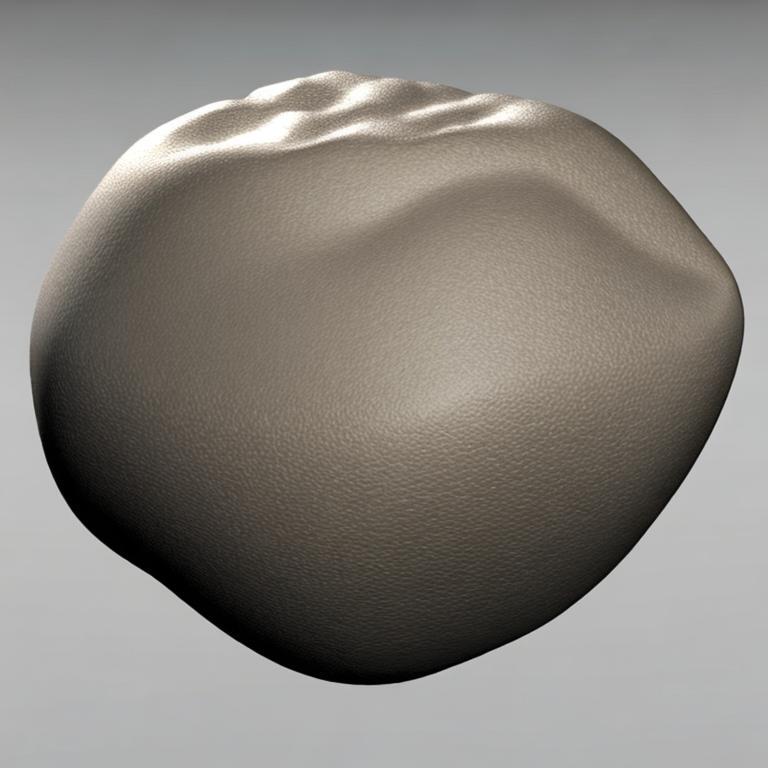}}{\emph{magama}\textsubscript{\ROUND}}
        \hspace{8pt}
        \jsubfig{\includegraphics[height=1.7cm]{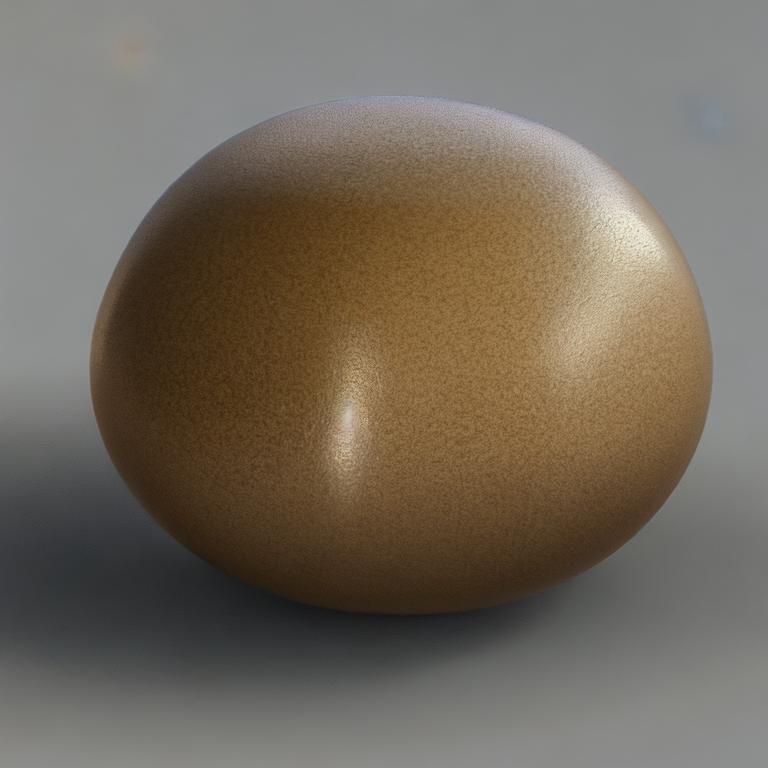}}{\emph{bomabo}\textsubscript{\ROUND}}
        \hspace{8pt}
        \jsubfig{\includegraphics[height=1.7cm]{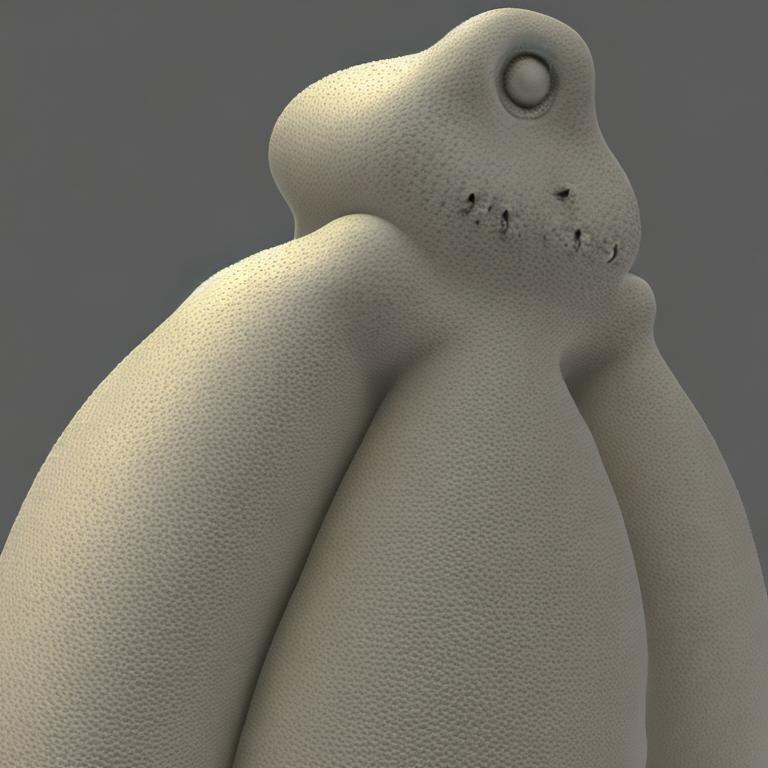}}{\emph{mamoma}\textsubscript{\ROUND}}
        \hspace{8pt}
        \jsubfig{\includegraphics[height=1.7cm]{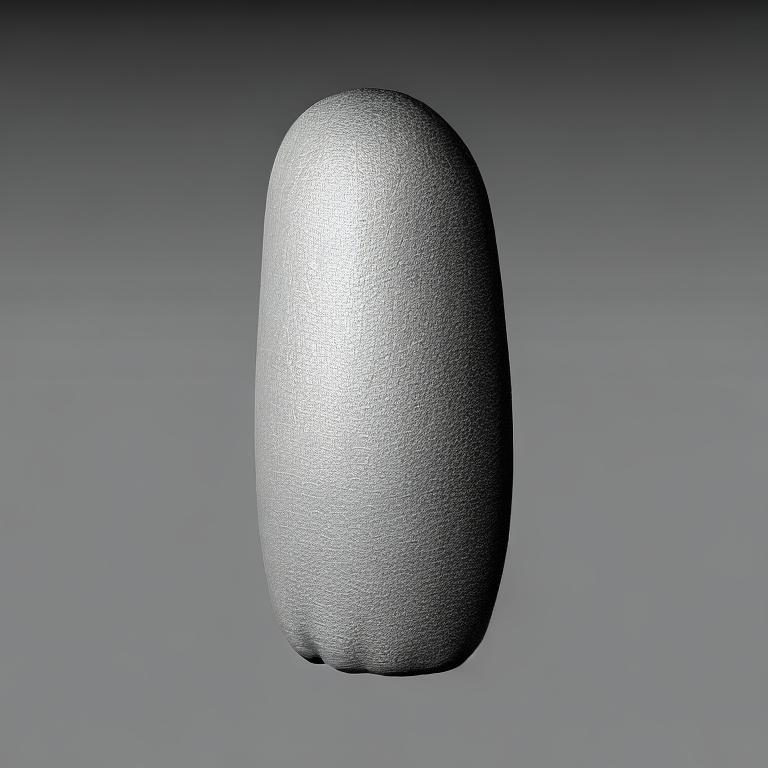}}{\emph{lobalo}\textsubscript{\ROUND}}
    \end{tabular}
    
    \caption{\textbf{Image generations for pseudowords with high (top 20\%) and low (bottom 20\%) geometric scores}. We visualize random selections of pseudoword--image pairs for each category. Pseudowords with class (\SHARP or \ROUND) that does not match its geometric score are indicated in \textcolor{red}{red}. As seen above, the shapes of the generated images noticeably correlate with the pseudoword class.
    }
    \label{fig:gen_examples}
\end{figure}

Qualitative results are shown in Table \ref{tab:sorted_real_words} and Figure \ref{fig:gen_examples}. In Table \ref{tab:sorted_real_words}, we see a striking pattern that the head and tail of this list represent adjectives which intuitively describe visual properties that are more ``sharp'' or ``round''. In other words, these properties are highly correlated with relative similarity to the two pseudoword classes \ROUND and \SHARP. This strengthens our quantitative results for ranking ground-truth adjectives with phonetic scores $\phi_{\text{\W}}$, as we see these scores are truly aligned with visual semantic properties for the VLMs under consideration (particularly for Stable Diffusion). 
The images in Figure \ref{fig:gen_examples} show a similarly striking pattern, with the sharper objects (as measured by $\gamma_{\text{\W}}$ score) being more commonly generated by pseudowords from \PseudoS and rounder objects by pseudowords from \PseudoR.

\begin{figure}
    \centering

    \begin{tabular}{c}
        \jsubfig{\includegraphics[height=1.6cm,cframe=white 1.5pt]{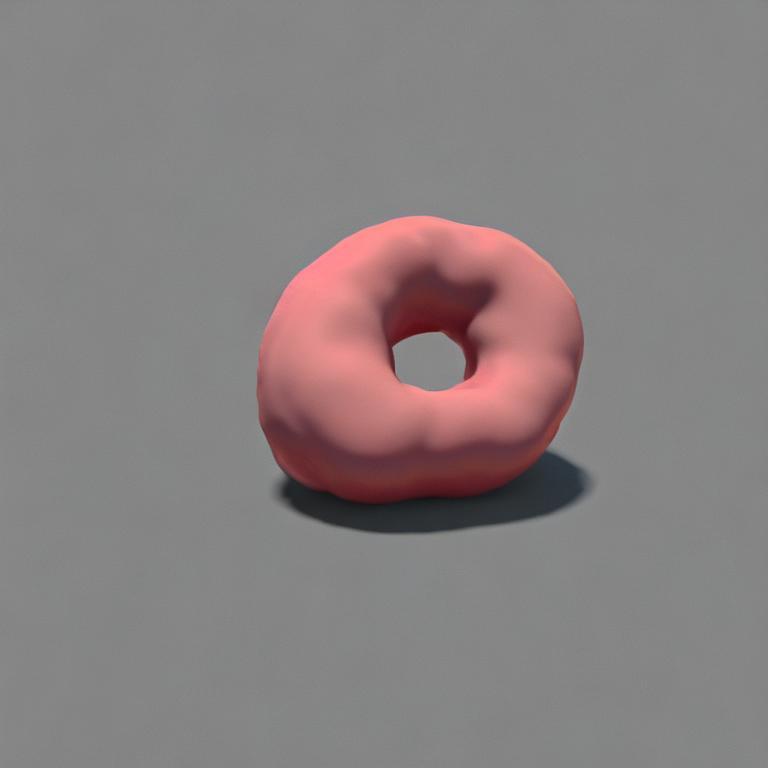}}{\emph{donudo}\textsubscript{\ROUND} (doughnut)}
        \hspace{3pt}
        \jsubfig{\includegraphics[height=1.6cm,cframe=white 1.5pt]{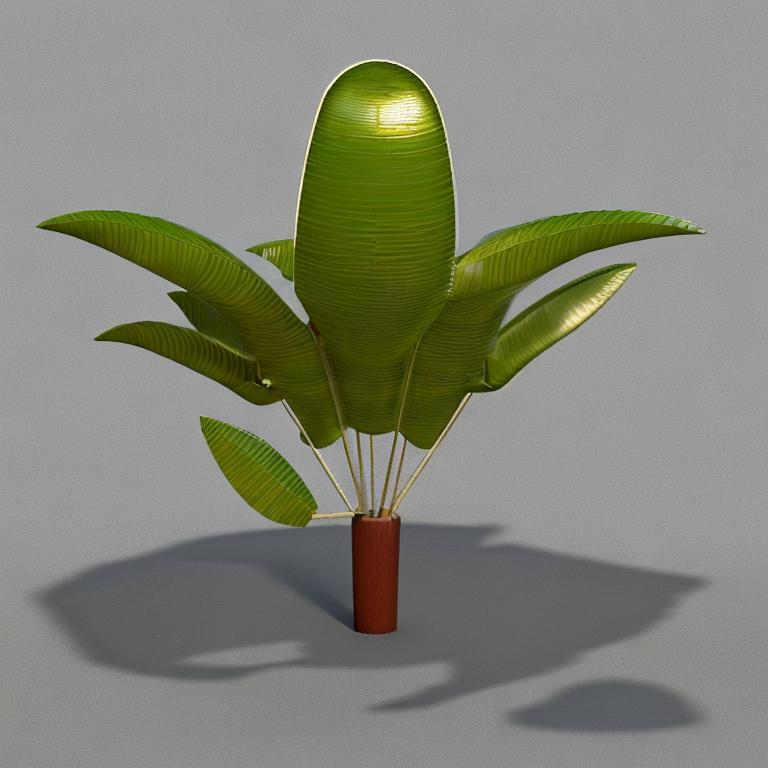}}{\emph{banaba}\textsubscript{\ROUND} (banana)}
        \hspace{3pt}
        \jsubfig{\includegraphics[height=1.6cm,cframe=white 1.5pt]{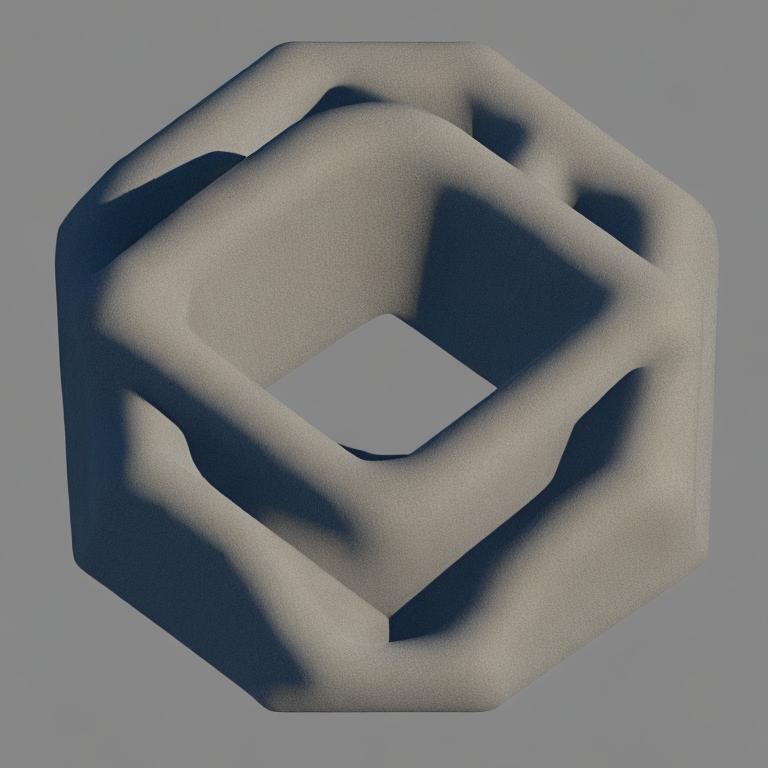}}{\emph{hexahe}\textsubscript{\SHARP} (hexagon)}
        \hspace{3pt}
        \jsubfig{\includegraphics[height=1.6cm,cframe=white 1.5pt]{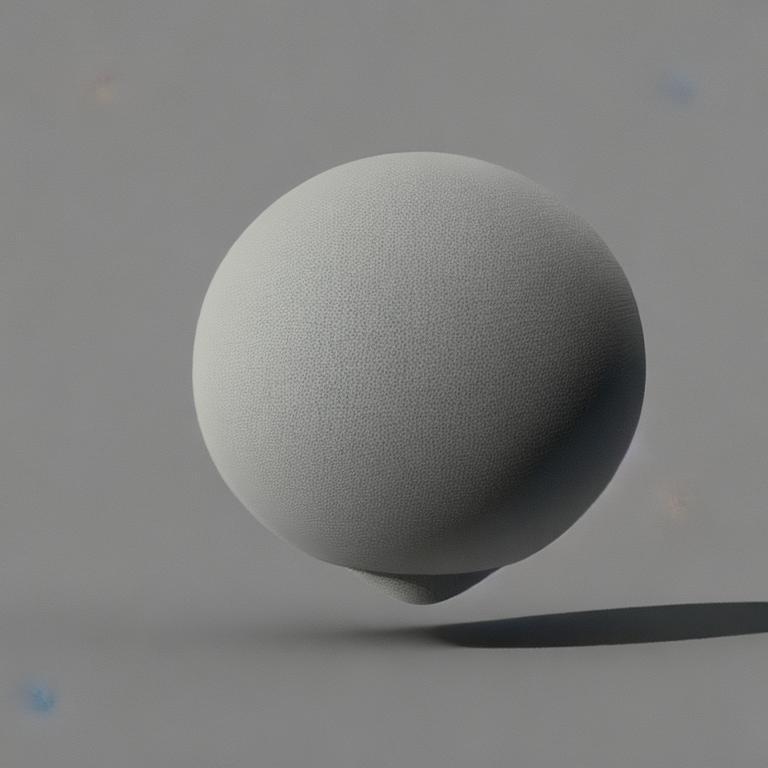}}{\emph{baloba}\textsubscript{\ROUND} (blob)}
        \hspace{3pt}
        \jsubfig{\includegraphics[height=1.6cm,cframe=white 1.5pt]{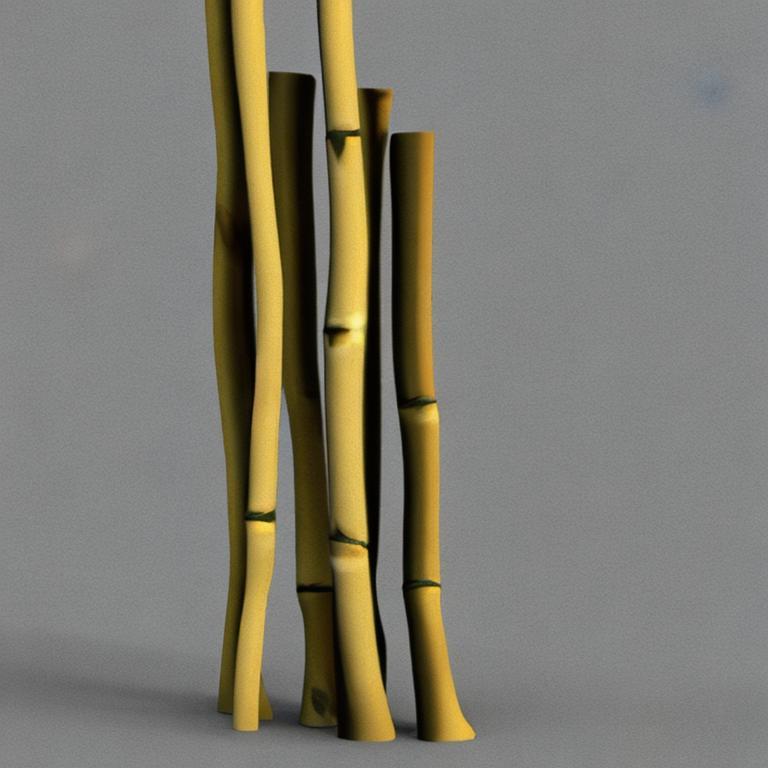}}{\emph{bamoba}\textsubscript{\ROUND} (bamboo)}
        \hspace{3pt}
        \jsubfig{\includegraphics[height=1.6cm,cframe=white 1.5pt]{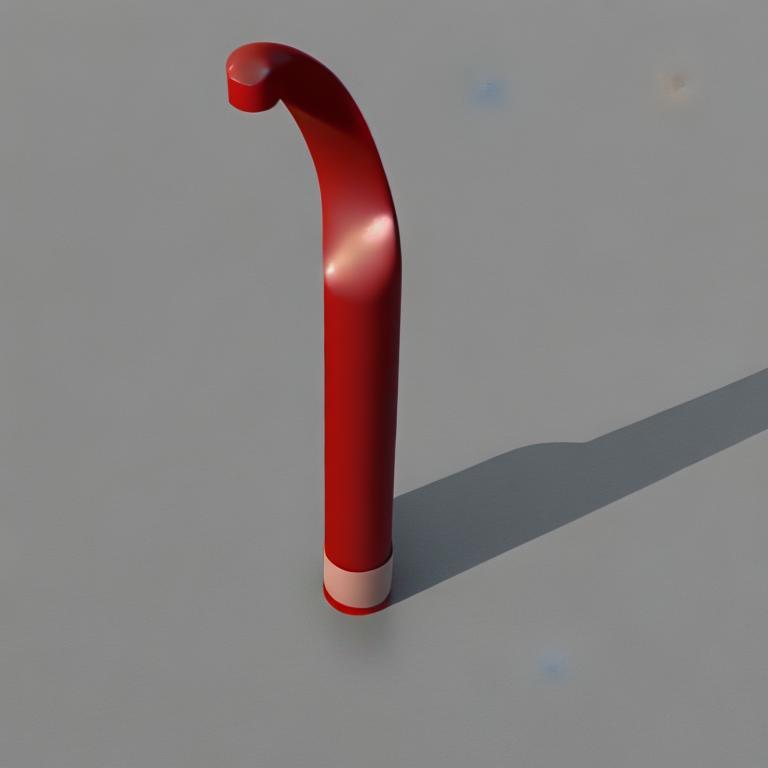}}{\textcolor{red}{\emph{pipepi}\textsubscript{\SHARP}} (pipe)}
        \hspace{3pt}
        \jsubfig{\includegraphics[height=1.6cm,cframe=white 1.5pt]{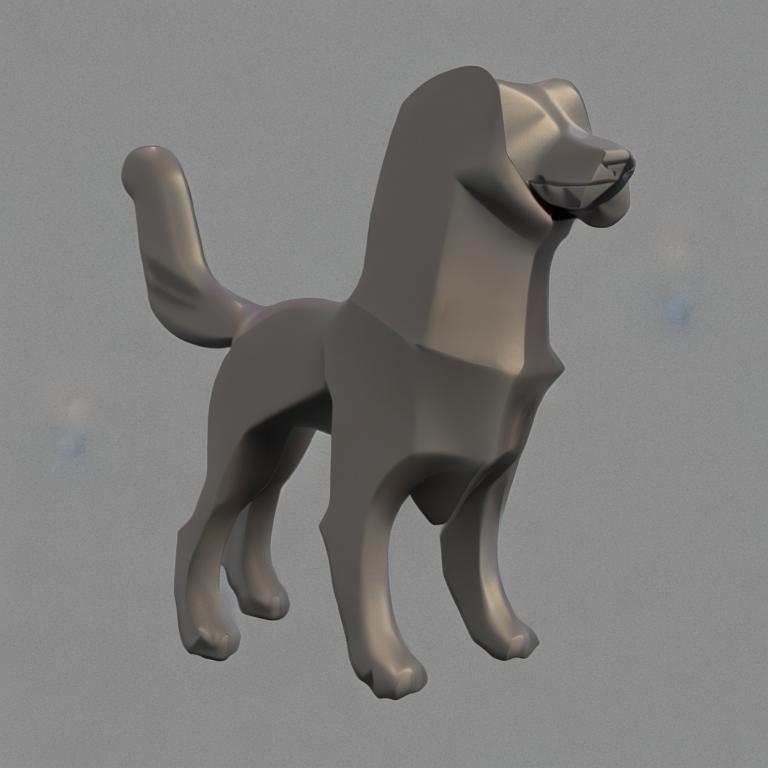}}{\textcolor{red}{\emph{dogado}\textsubscript{\ROUND}} (doggy)}
    \end{tabular}
    
    \caption{\textbf{Images generated from pseudowords reminiscent of real English words.} For each pseudoword we display an associated image generation and the automatically detected closest English word. %
    Pseudowords with high or low geometric scores (in top or bottom 20\% relative to all pseudowords) which do not match their class (\SHARP or \ROUND) are indicated in \textcolor{red}{red}.
    }
    \label{fig:real_word_examples}
\end{figure}

Pseudowords that resemble real English words may generate images reminiscent of real objects, as seen in Figure \ref{fig:real_word_examples}. There we display such pseudowords along with close English words, detected automatically via an automatic search heuristic (described in the supplementary material) combining fuzzy string matching and CLIP text-image similarity. This suggests a possible correlation between sounds in real English words and visual semantics as explored in the psycholinguistic literature~\cite{monaghan2014arbitrary,winter2021size,sidhu2021sound}. %

\section{Discussion, Limitations and Future Work}

We have shown that sound symbolism is reflected in VLMs by evaluating them on pseudowords with controlled phonetic properties. By comparing pseudowords built from known ``sharp'' or ``round'' speech sounds, we find that these models learn associations with corresponding sharp and round adjectival properties, parallel to the classic \emph{kiki--bouba} experiment in psychology. 

Further strengthening these findings, we provide strong evidence in the supplementary material that these models have not learned specifically from items illustrating the \emph{kiki--bouba} effect, by showing that this concept is not well-represented in the LAION dataset~\cite{schuhmann2022laion} upon which our VLMs were trained, and by showing that Stable Diffusion does not generate coherent content from prompts referring to the experiment itself. Rather, it appears that our VLMs have learned sound patterns from their training data via large-scale correlations between sounds and shapes. Future work could further explore complex generalization patterns and biases learned by VLMs, associating non-obvious visual or semantic meanings with particular input patterns.

Our findings have significance both in the field of multimodal machine learning as well as in psycholinguistics. From a computational perspective, our work reveals an emergent behavior in VLMs that has been hitherto unexplored, to the best of our knowledge. In general, VLMs are being used as black boxes without a full understanding of how they understand the visual semantics of language. Our work sheds light on how these models interpret and respond to language, as well as raising questions about how sound symbolic associations are inferred from our models' training datasets, whether additional dimensions of sound symbolic meaning are present in these models apart from our findings, and whether similar associations could exists with other modalities beyond vision.

From a cognitive science perspective, our work provides a new perspective to the extensive literature on sound symbolism, which has long been a topic of interest and debate. If VLMs learn sound symbolic association from valid text in caption data, this may be due to the presence of sound symbolism as reflected in the basic lexicon of English (or other languages), and the presence of sound symbolism in language itself is moderately controversial (notably denied by Ferdinand de Saussure as discussed in Section \ref{sec:intro}). In this vein, investigation into how VLMs infer sound symbolic associations from their training data could potentially shed light on how sound symbolism is learned during human language acquisition.
Additionally, these methods could provide new insights into the classic questions of what aspects of sound are tied to meaning and to what extent the observed phenomena are culturally universal or specific to the English language. Our work provides a new line of inquiry, suggesting that VLMs and their training datasets should be researched carefully for further insight.

Regarding limitations of our findings, we do not assume that these models directly imitate human cognition, although our results do show that sound symbolic patterns are present in captioned image data with a strong parallel to the psychology of language and vision. Additionally, we do not answer \emph{why} these associations are present in language or exactly how they are learned from the given data. In particular, we note that our results are agnostic to whether these models have memorized associations between particular letters and shapes, and to whether they have a deeper understanding of the phonetic or acoustic properties implied by these letters. We leave investigation of these questions to future work, and foresee them providing an important direction for research into sound symbolism in human cognition and machine learning.
\begin{ack}
This work was partially
supported by the Alon Fellowship. We thank Gal Fiebelman and Taelin Karidi for their helpful feedback.
\end{ack}

\AtNextBibliography{\small}
\printbibliography

\clearpage
\appendix
{\LARGE\textbf{Appendix}}

\section{Background on Phonetics and Sound Symbolism}

In this section, we provide background on phonetics of the English language, with an emphasis on the distinctions relevant to our work. Following the convention in modern linguistics, we use the International Phonetic Alphabet (IPA) to represent \emph{phonemes} (units of sound) and write IPA symbols between slashes \phon{\ldots}, while we write graphemic (written) representations in angle brackets \graphemic{\ldots}. Unless stated otherwise we use American English for transcriptions. For more thorough overviews of phonetics, the IPA, and the sounds used in English and across the languages of the world, see \cite{Ladefoged1995-sv,international1999handbook}.

In general, speech sounds may be roughly split into the categories of \emph{consonants} and \emph{vowels}, based on the degree of closure of the vocal tract.

For consonants, McCormick \emph{et al.}~\cite{mccormick_sound_2015} find that the most round-associated segments are all \emph{voiced} (sonorants \phon{m n l} and voiced stops \phon{b d g}). In this context, \emph{voicing} refers to vibration of the vocal cords during the production of a phoneme\footnote{Strictly speaking, phonemically ``voiced'' obstruents such as \phon{b} may be optionally devoiced in English making the relevant distinction one of aspiration, but this level of phonetic detail is not relevant for our discussion.}.
On the other end of the spectrum, they find a sharper association for the voiceless stop consonants \phon{p t k}, and for the \emph{fricative} and \emph{affricate} consonants \phon{f v s z tS dZ}, the latter produced by making a partial occlusion in the vocal tract to produce turbulent airflow.

Among vowels, McCormick \emph{et al.} find the dimension of \emph{roundedness} (in an articulatory sense) to be significant for sound symbolic association among vowels, with rounded vowels being more round-associated. In the context of articulatory phonetics, \emph{roundedness} refers to a vowel produced with lips pursed together. In English this is coupled with \emph{backness} which roughly refers to the position of the tongue relative to the back of the mouth; for example \phon{E} (as in \emph{bell}) is a front unrounded vowel while \phon{o} (as in \emph{bowl}) is a back rounded vowel.

Our split of English graphemes (letters) into \SHARP and \ROUND-associated categories for the purpose of constructing pseudowords is motivated by this phonetic background, along with a desire for balanced class sizes and avoiding digraphs (e.g. \graphemic{th}, which represents a single phoneme in English). For consonants, our \SHARP-associated graphemes \graphemic{p t k s h x} are all voiceless, fricatives or affricates while \ROUND-associated \graphemic{b d g m n l} are all voiced. For vowels, although English vowel orthography has a complex mapping to sound segments, we treat \graphemic{e i} as front unrounded vowels (and hence \SHARP) and \graphemic{o u} as back rounded vowels (and hence \ROUND) as a simplifying assumption, consistent with prior work. Since the grapheme \graphemic{a} can correspond to both front (\phon{\ae} as in \emph{fat}) and back (\phon{A} as in \emph{father}) vowels in English, we treat it as neutral with respect to sharpness associations\footnote{This is also consistent with previous work, which has used \graphemic{a} in both ``round'' stimuli like \emph{maluma} and ``sharp'' stimuli like \emph{takete}.}.

\section{Experimental Details}

\subsection{Image Generation Settings} \label{sec:imgen}

To generate images with Stable Diffusion, we use guidance scale 9 and 20 inference steps with DPM-Solver++~\cite{lu2022dpm,lu2022dpm2} as implemented in the Hugging Face \texttt{diffusers} library as \texttt{DPMSolverMultistepScheduler}. All images were generated using minibatches of 50 generated images for each prompt; 50 is chosen empirically based on the tradeoff between variance reduction (when averaging image embeddings) and computational constraints. All image generations use random seeds. In all figures shown in the main paper, the image displayed for each prompt is that closest to the centroid of the images in its respective minibatch (in OpenCLIP space, using embedding similarity).

\subsection{Model Checkpoints, Settings, and Compute}

We use the Hugging Face \texttt{transformers}~\cite{Wolf_Transformers_State-of-the-Art_Natural_2020} (v4.27.3) and \texttt{diffusers}~\cite{von_Platen_Diffusers_State-of-the-art_diffusion} (v0.14.0) APIs for loading the models under consideration from the following checkpoints:

\begin{itemize}
    \item OpenCLIP: \texttt{laion/CLIP-ViT-H-14-laion2B-s32B-b79K}
    \item Stable Diffusion: \texttt{stabilityai/stable-diffusion-2}
\end{itemize}

The only compute-intensive step in our method is producing image generations for every pseudoword. We use a single NVIDIA RTX A5000 GPU and are able to fit generation of a minibatch of 50 images in GPU memory by using attention slicing and VAE slicing, which reduce GPU memory consumption at the expense of inference speed. With these settings, generation uses approximately 23GB of GPU memory and generating a single minibatch takes about 6 minutes.

\subsection{Construction of Word Lists}

The lists of real English words used for qualitative evaluation---${\sim} 5.5K$ nouns denoted by N, and ${\sim}1K$ adjectives denoted by A---were constructed as follows. We first select all lemmas of synsets from WordNet~\cite{miller1995wordnet,miller1998wordnet} with noun and adjective part of speech labels respectively, along with the remaining adjectives from \AdjS and \AdjR. In order to avoid obscure and highly abstract words, we filter these lists using age of acquisition (AoA) and concreteness scores from the word lists of Kuperman \emph{et al.}~\cite{kuperman2012age} and Brysbaert \emph{et al.}~\cite{brysbaert2014concreteness}. In particular, we remove words with AoA $\geq 10$ and concreteness score $\leq 2.5$, along with words that do not appear in these datasets. We also manually remove a handful of obscenities and inappropriate items.

Examples of items from these lists include the following randomly-selected words:

\begin{tabular}{ll}
     N: & \emph{collarbone, poncho, lip, baseball, germ, swordsman, bumpiness, hitter, pilgrim} \\
     A: & \emph{electrical, shy, antiseptic, hearty, snappy, moist, lifeless, skinny, solitary, floury}
\end{tabular}

\subsection{Statistical Significance} \label{sec:supp_signif}

In Table \ref{tab:supp_signif}, we reproduce the Kendall correlation metrics from our main paper along with upper bounds on p-values. These are available since the Kendall correlation $\tau$ can be interpreted as a non-parametric test statistic, and we calculate these p-values using the \texttt{scipy.stats.kendalltau} function from the \texttt{scipy} library. We note that all of these results are statistically significant even at the very low significance threshold $\alpha = 10^{-3}$, indicating that these metric values would be highly unlikely under the null hypothesis in which there would be no underlying association between the quantities being compared. 

\begin{table}[t]
  \centering
  \setlength{\tabcolsep}{5.5pt}
  \def\arraystretch{0.95}
  \begin{tabularx}{0.56\columnwidth}{lcc}
    \toprule
    & $\gamma_{\text{\W}}$ & $\phi_{\text{\W}}$
    \\
    \cmidrule(lr){2-2}
    \cmidrule(lr){3-3}
    Model & $\tau$& $\tau$  \\
    \midrule
    Stable Diffusion & 0.34 ($p< 10^{-25}$) & 0.68 ($p < 10^{-3}$) \\
    CLIP & 0.39 ($p < 10^{-32}$) & 0.70 ($p < 10^{-3}$) \\
    \bottomrule
  \end{tabularx}
  \vspace{5pt}
  \caption{\textbf{Kendall correlation scores along with p-values} for models under consideration, demonstrating statistical significance. Interpretation of these p-values is discussed in Section \ref{sec:supp_signif}.}
\label{tab:supp_signif}
\end{table}
\section{Additional Results}

\subsection{Additional Multimodal Models} \label{sec:supp_add_models}

We provide results on additional text-conditioned image generation models, covering various model architectures, pretraining data and input languages. We consider the SOTA text-to-image diffusion models DeepFloyd-IF~\cite{deepfloyd} and Kandinsky~\cite{razzhigaev2023kandinsky}, as well as the SOTA text-conditioned GAN model GALIP~\cite{tao2023galip}. To evaluate these models, we use the methodology from our main paper. This includes prompts and evaluation metrics, calculated by embedding generated images with CLIP. We use the following model checkpoints from Hugging Face Model Hub to evaluate DeepFloyd-IF and Kandinsky: \texttt{DeepFloyd/IF-I-M-v1.0}, \texttt{kandinsky-community/kandinsky-2-1-prior}, \texttt{kandinsky-community/kandinsky-2-1}. For GALIP, we use the official code implementation with the checkpoint trained on the CC12M dataset.

We use the following inference settings: We run DeepFloyd-IF and Kandinsky in half-precision. For DeepFloyd-IF, we only run the first stage of inference which outputs low-resolution ($64 \times 64$) images. For Kandinsky, we use both prior and image-to-text pipelines with guidance scale set to $1.0$ and $768 \times 768$ output. All other inference settings use the defaults from the model checkpoint configuration files and the \texttt{DiffusionPipeline} class from the Hugging Face \texttt{diffusers} library. For GALIP, we use the default settings provided in the inference notebook in its official repository. For all of these models, we generate a single image for each pseudoword; for ``kiki'' and ``bouba'' generations to calculate $\Delta P_{kb}$, we generate 10 images for each for each.

In Table \ref{tab:supp_add_multimodal}, we show quantitative results for these models, which all show significant sound symbolic effects. We note that, in addition to architectural differences from Stable Diffusion, these models are all trained on different datasets -- in particular, GALIP was not trained on LAION data at all, precluding the data distribution in LAION as the only source of the observed effects.

\begin{table}[t]
  \centering
  \setlength{\tabcolsep}{5.5pt}
  \def\arraystretch{0.95}
  \begin{tabularx}{0.6\columnwidth}{lcccccc}
    \toprule
    &
    \multicolumn{3}{c}{$\gamma_{\text{\W}}$} & &
    \multicolumn{2}{c}{$\phi_{\text{\W}}$}
    \\
    \cmidrule(lr){2-4}
    \cmidrule(lr){6-7}
    \midrule
    Model & AUC & $\tau$ & $\Delta P_{kb}$ & & AUC & $\tau$  \\
    \midrule
    DeepFloyd-IF & 0.63 & 0.18 & 27\% & & 0.98 & 0.70 \\
    Kandinsky & 0.59 & 0.12 & 36\% & & 0.85 & 0.51 \\
    GALIP & 0.62 & 0.17 & 41\% & & 0.98 & 0.70 \\
    \midrule
    Stable Diffusion & 0.74 & 0.34 & 80\% & & 0.97 & 0.68 \\
    CLIP & 0.77 & 0.39 & 52\% & & 0.98 & 0.70 \\
    (random) & 0.50 & 0.00 & 0\% & & 0.50 & 0.00 \\
    \bottomrule
  \end{tabularx}
  \vspace{5pt}
  \caption{\textbf{Quantitative results for additional multimodal models.} We report results for additional SOTA text-to-image models with various architectures and pretraining data sources. We also report the results for the vision-and-language models considered in our work for comparison, as well as the random baseline indicated by (random).
  }
\label{tab:supp_add_multimodal}
\end{table}

\subsection{Unimodal Models}

As an additional comparison, we use our same methodology to probe unimodal text encoder models for sound symbolic associations. Since our metrics rely on cosine similarity of pooled embeddings, we use Sentence Transformer models~\cite{reimers-2019-sentence-bert} which were trained with a cosine similarity-based contrastive semantic objective. In particular, we compare the following models:

\begin{itemize}
\item SMPNet: MPNet~\cite{song2020mpnet} fine-tuned with a Sentence Transformer objective (Hugging Face checkpoint \texttt{sentence-transformers/all-mpnet-base-v2}, 109M parameters)
\item SDRoBERTa: DistilRoBERTa~\cite{Sanh2019DistilBERTAD} fine-tuned with a Sentence Transformer objective (Hugging Face checkpoint \texttt{sentence-transformers/all-mpnet-base-v2}, 82M parameters)
\item SMiniLM: MiniLM~\cite{wang2020minilm} fine-tuned with a Sentence Transformer objective (Hugging Face checkpoint \texttt{sentence-transformers/all-MiniLM-L12-v2}, 33M parameters)
\item SALBERT: ALBERT~\cite{lan2019albert} fine-tuned with a Sentence Transformer objective (Hugging Face checkpoint \texttt{sentence-transformers/paraphrase-albert-small-v2}, 2M parameters)
\end{itemize}

Quantitative results are shown in Table \ref{tab:supp_unimodal_metrics}. Interestingly, the larger unimodal models SMPNet and SDRoBERTa show stronger sound symbolic associations, while the smaller SMiniLM and SALBERT models show near-random performance by most metrics. We also show qualitative results for SMPNet, the unimodal model with overall highest sound symbolic metrics, in Table \ref{tab:supp_unimodal_sorted}. There it can be seen that the model appears to partially reflect ``sharp'' and ``round'' semantic associations using our phonetic scoring method, but also partially reflects the surface spelling of the English words being evaluated. These mixed results raise questions regarding the ability of unimodal models to learn sound symbolic associations from text data alone and the relative contributions of textual and image data to this phenomenon; we leave a thorough investigation of these topics to future work.

\begin{table}[t]
  \centering
  \setlength{\tabcolsep}{5.5pt}
  \def\arraystretch{0.95}
  \begin{tabularx}{0.6\columnwidth}{lcccccc}
    \toprule
    &
    \multicolumn{3}{c}{$\gamma_{\text{\W}}$} & &
    \multicolumn{2}{c}{$\phi_{\text{\W}}$}
    \\
    \cmidrule(lr){2-4}
    \cmidrule(lr){6-7}
    \midrule
    Model & AUC & $\tau$ & $\Delta P_{kb}$ & & AUC & $\tau$  \\
    \midrule
    SMPNet & 0.76 & 0.37 & 42\% & & 0.82 & 0.46 \\
    SDRoBERTa & 0.74 & 0.34 & 62\% & & 0.71 & 0.30 \\
    SMiniLM & 0.53 & 0.04 & 60\% & & 0.57 & 0.10 \\
    SALBERT & 0.50 & 0.01 & 44\% & & 0.54 & 0.06 \\
    \midrule
    Stable Diffusion & 0.74 & 0.34 & 80\% & & 0.97 & 0.68 \\
    CLIP & 0.77 & 0.39 & 52\% & & 0.98 & 0.70 \\
    (random) & 0.50 & 0.00 & 0\% & & 0.50 & 0.00 \\
    \bottomrule
  \end{tabularx}
  \vspace{5pt}
  \caption{\textbf{Quantitative results for unimodal models.} We calculate the metrics used in our main paper for four different unimodally (text-only) trained models, all fine-tuned with the Sentence Transformer objective which makes cosine similarity-based probing semantically meaningful for text encoders. We also report the results for the vision-and-language models considered in our work for comparison, as well as the random baseline indicated by (random).
  }
\label{tab:supp_unimodal_metrics}
\end{table}

\begin{table}[t]
  \centering
  \setlength{\tabcolsep}{5.5pt}
  \def\arraystretch{0.95}
  \begin{tabularx}{0.99\columnwidth}{lll}
    \toprule
    POS & Lowest $\phi_{\text{\W}}$ ($\Rightarrow$ \ROUND) & Highest $\phi_{\text{\W}}$ ($\Rightarrow$ \SHARP)\\
    \midrule
    \multicolumn{3}{l}{\textbf{SMPNet}} \\
     Noun & \literal{baboon, mongoose, moo, hobo, bogeyman,} & \literal{hexagon, tyke, skateboarder, pillar, pinstripe, hive,} \\
    & \literal{gab, bedbug, loon, gremlin, ladybug, noodle} & \literal{yoke, pinecone, teakettle, pane, whiskers, star} \\
    \rule{0pt}{3ex}
    Adj. & \literal{grizzly, lunar, beaded, moonless, bluish,} & \literal{khaki, shipshape, teensy, twinkly, bladed, starlit,} \\
    & \literal{muddy, gooey, wormy, moonlit, blubbery} & \literal{pointed, teenage, whiskered, scaly, spiky} \\
    \bottomrule
  \end{tabularx}
  \vspace{5pt}
  \caption{\textbf{Real English words sorted by phonetic score for SMPNet,} the unimodal model under consideration with highest overall sound symbolic metrics. As seen above, the order of words partially matches intuition about ``round'' and ``sharp'' characteristics, but also appears to moderately correlate with the initial letters in the English words (e.g. many words in the left column begin with \ROUND-associated letters \emph{b-}. \emph{m-}, \emph{g-}, or \emph{l-}).}
\label{tab:supp_unimodal_sorted}
\end{table}

\subsection{Additional Prompts}

We provide a comparison of results using various prompts with CLIP in Table \ref{tab:supp_more_prompts}, with the prompt used in our main paper (\emph{a 3D rendering of a \W object}) in the final row. In all cases, we use the prompt as-is when inserting adjectives, and with the word ``shaped'' added (e.g. \emph{a 3D rendering of a \W shaped object}) when inserting nouns and pseudowords. As can be seen there, all prompts under consideration display sound symbolic effects despite some variation in metric values.

\begin{table}[t]
  \centering
  \setlength{\tabcolsep}{5.5pt}
  \def\arraystretch{0.95}
  \begin{tabularx}{0.75\columnwidth}{lcccccc}
    \toprule
    &
    \multicolumn{3}{c}{$\gamma_{\text{\W}}$} & &
    \multicolumn{2}{c}{$\phi_{\text{\W}}$}
    \\
    \cmidrule(lr){2-4}
    \cmidrule(lr){6-7}
    \midrule
    Prompt & AUC & $\tau$ & $\Delta P_{kb}$ & & AUC & $\tau$  \\
    \midrule
    \emph{a \W object} & 0.78 & 0.40 & 32\% & & 0.96 & 0.67 \\
    \emph{a picture of a \W object} & 0.83 & 0.46 & 67\% & & 0.99 & 0.71 \\
    \emph{an oil painting of a \W object} & 0.75 & 0.35 & 32\% & & 1.00 & 0.73 \\
    \emph{a \W thing} & 0.82 & 0.45 & 35\% & & 0.98 & 0.70 \\
    \emph{a \W item} & 0.75 & 0.35 & -35\% & & 0.92 & 0.61 \\
    \emph{a \W drawing} & 0.84 & 0.48 & 17\% & & 0.95 & 0.65 \\
    \emph{this thing is \W} & 0.79 & 0.41 & 59\% & & 0.91 & 0.59 \\
    \emph{\W} & 0.77 & 0.39 & 18\% & & 0.93 & 0.62 \\
    \midrule
    \emph{a 3D rendering of a \W object} & 0.77 & 0.39 & 52\% & & 0.98 & 0.70 \\
    \bottomrule
  \end{tabularx}
  \vspace{5pt}
  \caption{\textbf{Results on various prompts}, evaluated using CLIP with our probing methods. The prompt used in our main paper is reported in the last row. Although the metric values vary somewhat by prompt, all of the prompts under consideration exhibit sound symbolic effects.}
\label{tab:supp_more_prompts}
\end{table}

\subsection{Multilingual Results}

To investigate whether sound symbolism may exist in a multilingual VLM, we evaluate the Kandinsky~\cite{razzhigaev2023kandinsky} multilingual text-to-image model with our methodology on prompts in multiple languages. While we evaluate Kandinsky on English prompts in Section \ref{sec:supp_add_models}, this model can also receive multilingual prompts (and uses multilingual CLIP as its text encoder). In particular, we construct prompts (shown in Table \ref{tab:supp_multiling_prompts}) in four geographically and linguistically diverse languages: Finnish, Indonesian, Hungarian, and Lithuanian. Using the same inference methodology as in Section \ref{sec:supp_add_models}, we evaluate Kandinsky on these prompts; results are shown in Table \ref{tab:supp_multiling_results}.  We find non-trivial sound symbolism in this setting in each language, suggesting that sound symbolism may be learned in a multilingual vision-and-language setting.

We place these results in context by emphasizing that we do not claim to demonstrate the universality of sound symbolism across languages; rather, our work work focuses on showing the that common VLMs are observed to learn sound symbolic associations. Nevertheless, these positive results in a multilingual setting are suggestive and further investigation of multilingual vision-and-language models might provide insight into cross-linguistic sound symbolic patterns in language.

\begin{table}[t]
  \centering
  \setlength{\tabcolsep}{5.5pt}
  \def\arraystretch{0.95}
  \begin{tabularx}{0.67\columnwidth}{ll}
    \toprule
    Language & Prompt  \\
    \midrule
    Finnish & \emph{3D-renderöinti objektista, jolla on muoto: ``\W''} \\
    Indonesian & \emph{rendering 3D objek dengan bentuk: ``\W''} \\
    Hungarian & \emph{egy objektum 3D-s megjelenítése alakzattal: ``\W''} \\
    Lithuanian & \emph{3D objekto atvaizdavimas su forma: ``\W''} \\
    \bottomrule
  \end{tabularx}
  \vspace{5pt}
  \caption{\textbf{Multilingual prompts} used for results in Table \ref{tab:supp_multiling_results}. \W indicates the slot where pseudowords are inserted. Each prompt roughly translates to English \emph{a 3D rendering of an object with shape: ``\W''}, chosen for cross-linguistic grammatical uniformity.
  }
  \label{tab:supp_multiling_prompts}
\end{table}
\begin{table}[t]
  \centering
  \setlength{\tabcolsep}{5.5pt}
  \def\arraystretch{0.95}
  \begin{tabularx}{0.55\columnwidth}{lcccccc}
    \toprule
    &
    \multicolumn{3}{c}{$\gamma_{\text{\W}}$} & &
    \multicolumn{2}{c}{$\phi_{\text{\W}}$}
    \\
    \cmidrule(lr){2-4}
    \cmidrule(lr){6-7}
    Language & AUC & $\tau$ & $\Delta P_{kb}$ & & AUC & $\tau$  \\
    \midrule
    Finnish & 0.69 & 0.27 & -4\% & &	0.94 & 0.64 \\
    Indonesian & 0.67 & 0.24 & 4\% & &	0.94 & 0.64 \\
    Hungarian  & 0.60 & 0.14 & 9\% & & 0.93 & 0.62 \\
    Lithuanian  & 0.73 & 0.32 & 23\% & & 0.97 & 0.68 \\
    \midrule
    (random) & 0.50 & 0.00 & 0\% & & 0.50 & 0.00 \\
    \bottomrule
  \end{tabularx}
  \vspace{5pt}
  \caption{\textbf{Multilingual results} for generations, using the Kandinsky text-to-image model with prompts from Table \ref{tab:supp_multiling_prompts}. (random) indicates the expected performance of a purely random scoring method. Recall that the AUC and $\tau$ metrics are calculated over our full set of 648 pseudowords, and $\Delta P_{kb}$ is calculated using ``kiki'' and ``bouba'' alone.
  }
   \label{tab:supp_multiling_results}
\end{table}

\subsection{Corner Detection Probing}

To provide an additional visually grounded probe for sound symbolism in text-to-image models, we estimate the sharpness or roundness of images with a corner detection algorithm (as more corners generally correspond to a visually more ``sharp'' image). In particular, we apply the Harris corner detector~\cite{harris1988combined} to our pseudoword image generations produced by Stable Diffusion. We use the \texttt{cornerHarris} implementation in the \texttt{OpenCV} library with parameters $(5, 15, 0.04)$ to our images (of size $768 \times 768$) and set negative output values to zero. We then apply non-maximum suppression using a $100 \times 100$ sliding window. Finally, we calculate the maximum output value $M$ and count the number of outputs that are greater than $0.01 * M$. This yields the number of corner detections in the given image.

Applying this to Stable Diffusion generations for pseudowords, we find that ``round'' and ``sharp'' pseudowords correspond to $11.47$ and $12.75$ corners per image on average, respectively. To determine the significance of this effect, we apply a two-sided Welch's t-test to the number of corners per image for the 32.4K image generations (50 per pseudoword), testing for a significant difference in the mean number of corners per image between the ``sharp'' and ``round'' association classes (each comprising exactly half of the items). This test yields t-statistic $12.028$ with $p < 10^{-32}$, indicating a significant difference between the means of the two classes. In other words, we find a significant difference in the number of corner detections for images generated from ``sharp'' and ``round'' pseudowords, confirming their visual distinctness on average.

\subsection{Kiki--Bouba in LAION}

\begin{figure}
    \centering

    \begin{tabular}{c}
        \jsubfig{\includegraphics[height=6.5cm]{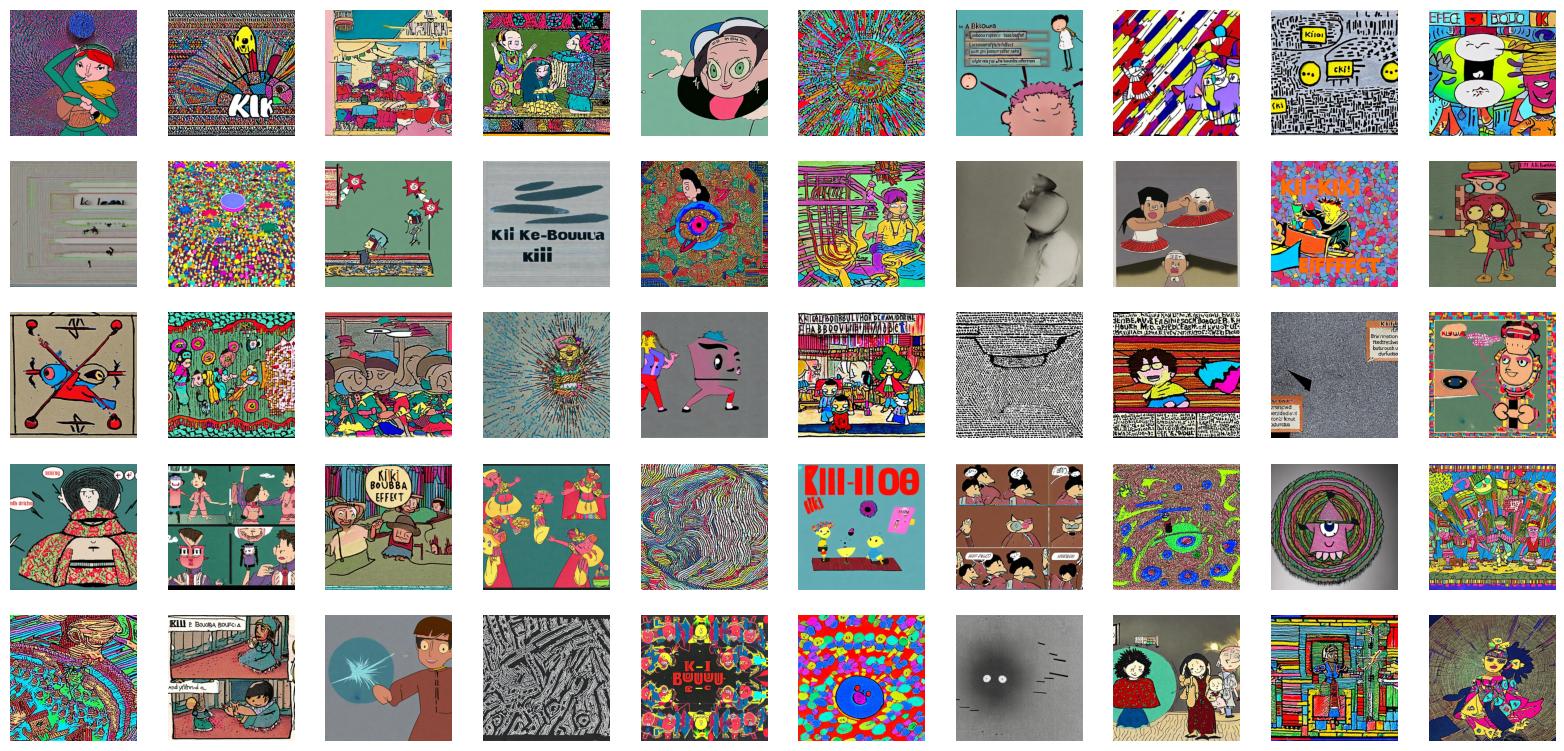}}{\emph{``a picture of the kiki-bouba effect''}} \\
        \\
        \jsubfig{\includegraphics[height=6.5cm]{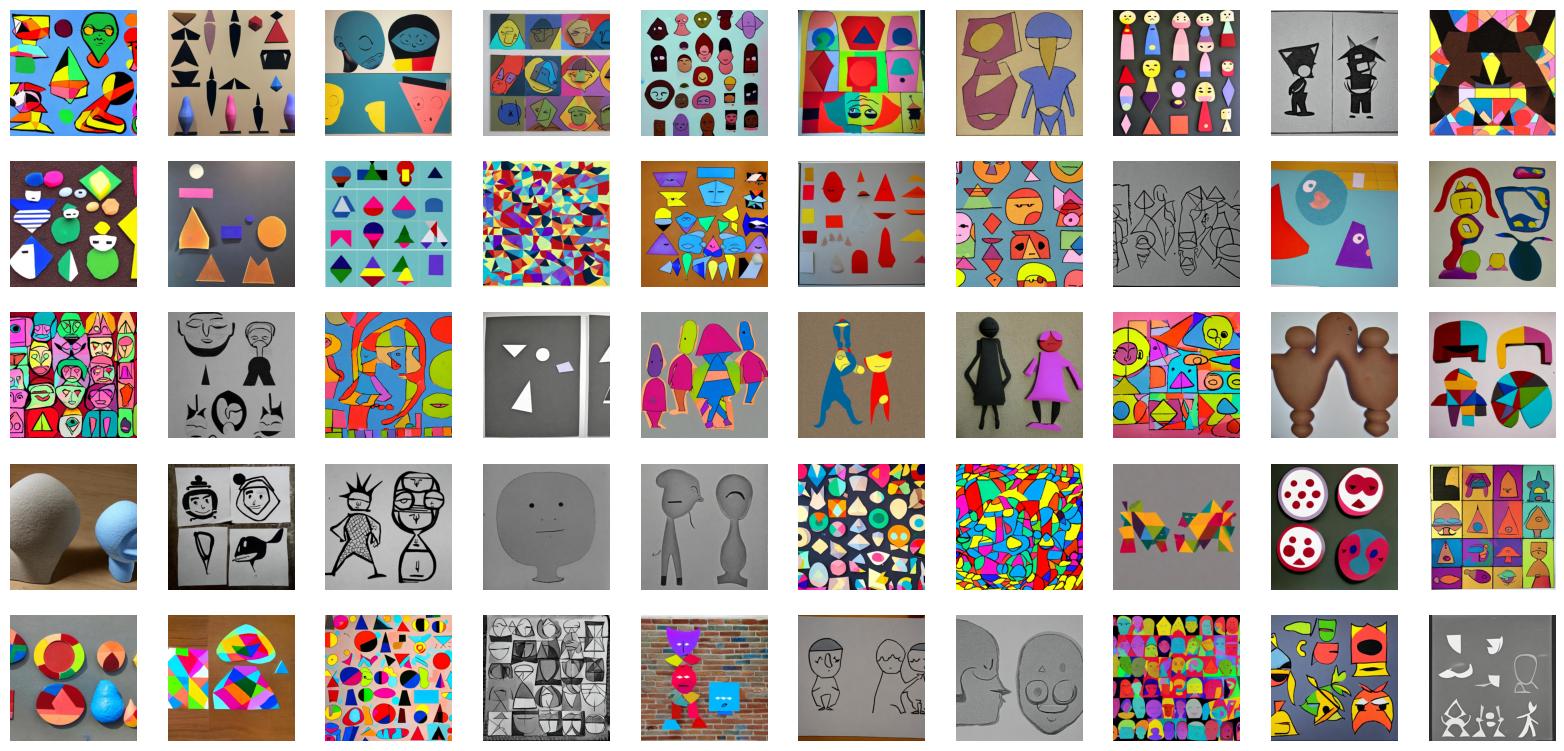}}{\emph{``the shapes bouba and kiki as used in psychological research''}} \\
        \\
        \jsubfig{\includegraphics[height=6.5cm]{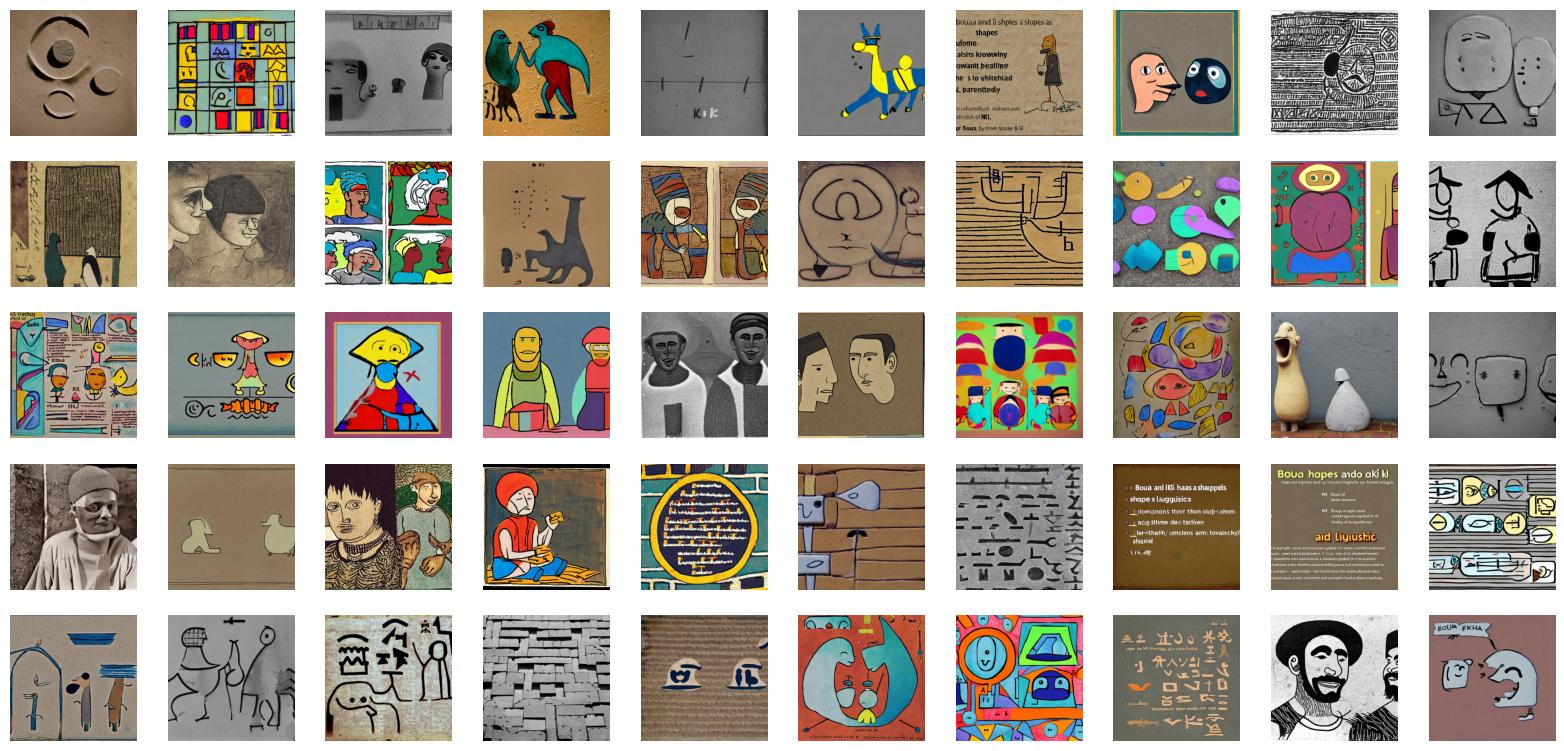}}{\emph{``bouba and kiki shapes as known from linguistics''}} \\
    \end{tabular}
    
    \caption{\textbf{Image generations for prompts describing the kiki--bouba effect.}. We display all images from a single (size 50) minibatch of text to image generations for each of the given prompts describing the \emph{kiki--bouba} effect. The generations do not coherently depict the psychological effect as a concept.}
    \label{fig:supp_kb_experiment_prompts}
\end{figure}

We first manually examine image generations for various prompts that directly refer to the \emph{kiki--bouba} effect, to evaluate whether Stable Diffusion (trained on the LAION dataset) shows signs of recognizing this as a concept. We generate minibatches (50 images each) of images for a variety of such prompts, listed below. From visual inspection, none of these prompts yield images with a coherent relation to the \emph{kiki--bouba} effect. Examples of full minibatches of image generations for some of these prompts are shown in Figure \ref{fig:supp_kb_experiment_prompts}.

Prompts manually examined include: \emph{a picture of the kiki-bouba experiment, a picture of the kiki/bouba experiment, a picture of the bouba-kiki experiment, a picture of the bouba/kiki experiment, a picture of kiki and bouba, a picture of bouba and kiki, kiki and bouba, bouba and kiki, a picture of the kiki-bouba effect, a picture of the kiki/bouba effect, a picture of the bouba-kiki effect, a picture of the bouba/kiki effect, shapes for the bouba-kiki experiment, shapes for the bouba-kiki effect, shapes for the kiki-bouba experiment, shapes for the kiki-bouba effect, bouba and kiki psychological stimuli, the shapes bouba and kiki as used in psychological research, the bouba/kiki stimuli in psychology, bouba and kiki shapes as known from linguistics}

These results suggest that this concept is not readily learnable from the LAION dataset; to further strengthen this hypothesis, we perform a search of the LAION dataset, using the ${\sim}2B$ LAION-2B subset of primarily English captions available on the Hugging Face datasets hub as \texttt{laion/laion2B-en}\footnote{\url{https://huggingface.co/datasets/laion/laion2B-en}}. We use caption data alone (without the corresponding images) and only consider items indicated as non-NSFW in the accompanying metadata; we preprocess by converting captions to lowercase. Searching for captions containing both \emph{kiki} and \emph{bouba} as substrings and excluding those containing the West African name \emph{Boubacar}, we are left with only the following 12 captions in the entire dataset:

\begin{itemize}
    \item \emph{the bouba kiki effect and language are there certain human sounds with meanings that can cross the language  in the bouba-kiki effect,  and that you want to know if the picture is bouba or kiki.}

\item \emph{kiki or bouba: what is the shape of your taste?}

\item \emph{ramachandran and hubbard [3] suggest that the kiki/bouba effect has implications for the evolution of language, because it suggests that the naming of}

\item \emph{preschool-age sound- shape correspondences to the bouba-kiki effect karlee jones, b.s. ed. \& matthew carter, ph.d. valdosta state university.}

\item \emph{the bouba-kiki effect}

\item \emph{the bouba kiki effect and language if you were looking at two shapes—specifically, a pointy, jagged polygon and an amoeboid-like splotch—which would you name bouba, and which would you name kiki.}

\item \emph{grooving: kiki and bouba minds}

\item \emph{bumblebee and how to design the transformers || the kiki/bouba effect}

\item \emph{the bouba/kiki effect recent work by daphne maurer and colleagues has shown that even children as young as 2.5 (too young to read) show this effect.}

\item \emph{the bouba / kiki effect}

\item \emph{the bouba kiki effect and language if you were looking at two shapes—specifically, a pointy, jagged polygon and an amoeboid-like splotch—which would you name bouba, and which would you name kiki.}

\item \emph{the bouba/kiki effect the bouba/kiki effect was first observed by german-american psychologist wolfgang köhler in 1929.}
\end{itemize}

The learnability of rare concepts for text-to-image generation has been explicitly studied by Samuel \emph{et al.}~\cite{samuel2023all}, who find that Stable Diffusion struggles to accurately depict concepts with less than 10,000 samples in LAION-2B. Therefore we conclude that it is highly unlikely that Stable Diffusion has direct knowledge of the \emph{kiki--bouba} effect as an abstract concept.

\subsection{Pseudowords Resembling Real English Words}\label{sec:eng_sim}

In order to automatically find the English word of which a given pseudoword is reminiscent (e.g. \emph{donudo} -- ``doughnut''), we use the following heuristic algorithm. Given English word \W and pseudoword \PW, we calculate:

\begin{enumerate}
\item $s_{clip}(\text{\W}, \text{\PW})$: The cosine similarity of CLIP embeddings of the word \W (as text) and the mean embedding of images generated using the pseudoword \PW.
\item $s_{text}(\text{\W}, \text{\PW})$: Text similarity score ($\in [0, 1]$) based on Levenshtein edit distance~\cite{navarro2001guided}. In particular, let $\left|\text{\W}\right|$ and $\left|\text{\PW}\right|$ be the length in characters of \W and \PW respectively, $d$ the edit distance between \W and \PW, and $m = \max\{\left|\text{\W}\right|, \left|\text{\PW}\right|\}$; then we define $s_{text} := 1 - d / m$.
\end{enumerate}

We define the score $s(\text{\W}, \text{\PW}) := s_{clip}(\text{\W}, \text{\PW}) \cdot s_{text}(\text{\W}, \text{\PW})$. Intuitively, this is maximized when \W is both textually similar to \PW and semantically matches the images generated from \PW. On input \PW, our heuristic returns $\arg\max_{\text{\W}}s(\text{\W}, \text{\PW})$, the English word maximizing this score. We constrain this search to basic, concrete English words by filtering with age of acquisition (AoA) and concreteness scores found in the word lists of Kuperman \emph{et al.}~\cite{kuperman2012age} and Brysbaert \emph{et al.}~\cite{brysbaert2014concreteness}; in particular, we use the ${\sim 8.3K}$ English words from these lists with AoA $ < 10$ and concreteness $< 2.5$.
\section{User Study Details}

\subsection{IRB Approval, Participant Sourcing, and Compensation}

Our user study, which received approval from our institution's IRB, was conducted using the Amazon Mechanical Turk (MTurk) crowdsourcing platform. Our surveys were made available to MTurk workers with at least 1000 completed HITs (MTurk tasks) and a HIT approval rate of at least 95\%. Workers accepted a consent statement (reproduced below) in order to proceed, including confirmation of being age 18 or above. Workers were fully anonymized other than collecting their MTurk worker IDs, a unique non-identifiable code associated with each worker. Workers were compensated according to the length of each survey: the shorter 15-minute survey (version A; \emph{kiki-bouba}) paid \$2.50 upon completion and the longer 25-minute survey (version B; random pseudowords) paid \$4.25, as seen when accepting the task.

\subsection{Consent Statement}

To participate in the survey, workers were required to read and agree to the following consent statement (contact information has been redacted but was visible to workers; the listed time depended on the survey version):

\emph{Consent to Participate in Online Survey Research Using MTurk}

\emph{Study Description:  We are researchers at the Tel Aviv University doing a research study about cognition, vision and language. If you agree to participate, you will be asked to complete an online survey that will take approximately (time) minutes to complete.}

\emph{Risks/ Benefits:  Risks to participants are considered minimal.  Collection of data and survey responses using the internet involves the same risks that a person would encounter in everyday use of the internet, such as fatigue or breach of confidentiality.  While the researchers have taken every reasonable step to protect your confidentiality, there is always the possibility of interception or hacking of the data by third parties that is not under the control of the research team. There will be no costs for participating. Benefits of participating include payment from Amazon as described in the HIT description.}

\emph{Confidentiality: Researchers will have access to your MTurk worker ID which may be able to link to your personal information on your Amazon public profile page, depending on your settings you have on your Amazon profile.  MTurk worker IDs will not be shared with anyone outside the study team and will be used solely for the purposes of distributing compensation and will not be stored with your responses. We will not be accessing any personally identifying information about you that you may have put on your Amazon public profile page. Any reports and presentations about the findings from this study will not include your name or any other information that could identify you.  }

\emph{Voluntary Participation:  Your participation in this study is voluntary.  You may choose to not answer any of the questions or withdraw from this study at any time without penalty.  Your decision will not change any present or future relationship with the Tel Aviv University or Amazon.}

\emph{For more information about the study or study procedures, contact (redacted).}

\emph{Research Subject’s Consent to Participate in Research: By entering this survey/ selecting I agree, you are indicating that you have read the consent form, you are age 18 or older and that you voluntarily agree to participate in this online research study. Please make sure that you have read and agree to Amazon’s Mechanical Turk participant and privacy agreements as these may impact the disclosure and use of your personal information.}

\subsection{Survey Instructions}

\begin{figure}
    \centering
    \includegraphics[width=1\textwidth]{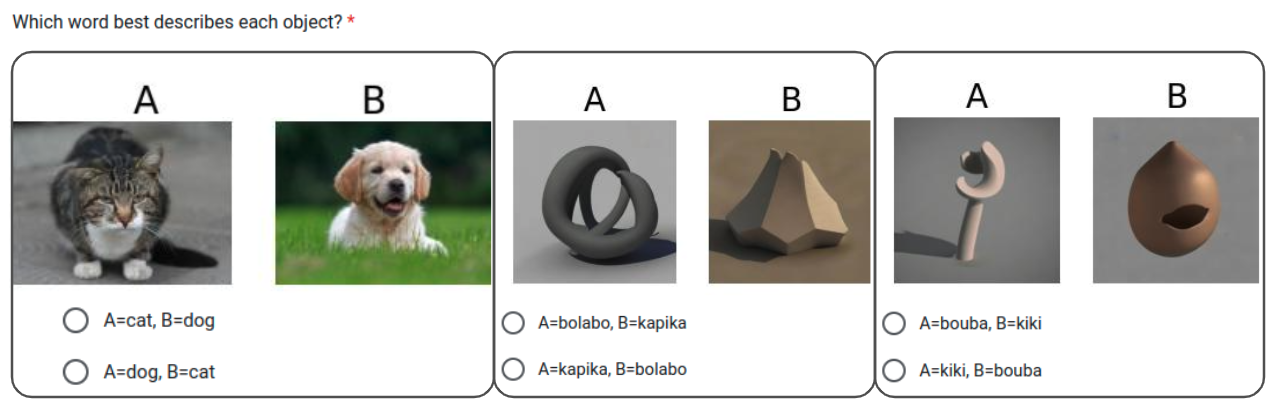}
    
    \caption{\textbf{Screenshots of questions from our user study.} The left-hand screenshot shows the sample question shown to all participants at the beginning of the survey. The screenshot in the center shows a question from the pseudowords version of the survey, while the right-hand screenshot shows a question from the \emph{kiki--bouba} version. }
    \label{fig:supp_survey}
\end{figure}

Participants in our user study were shown the following instructions describing the task to complete:

\emph{In the following section you will be shown pairs of images, labelled as "A" and "B" and showing two objects.
For each image pair, you will be asked which object is better described by different made-up words.
Choose whichever feels more appropriate for the given images, even if you are not sure. 
For example, in the question below you should pick "A=cat, B=dog".}

Underneath this appeared the sample image pair and options shown on the left of Figure \ref{fig:supp_survey}. Upon proceeding, they were given the full page of 20 questions beginning with the following instruction:

\emph{You will now see 20 image pairs and will be asked which nonsense words best match each image.
Pick whichever feels most right to you, even if you are not sure.}

The 20 questions following this were in the format illustrated in Figure \ref{fig:supp_survey}, either as in the center or the right-hand side of the figure depending on the survey version. The order that questions appeared was fully randomized for each participant. All of these questions were forced-choice, and participants could only submit after answering all of them. 

Finally, participants were asked the (required) question

\emph{Had you heard of the "kiki-bouba effect" before taking this survey?}

along with an optional field for comments.

\subsection{Survey Design Methodology}\label{sec:design}

We distributed two primary versions of our survey separately: version A, using images generated with prompts using \emph{kiki} or \emph{bouba} (and asking participants to choose between these words); and version B, using images generated with pseudowords from \PseudoS and \PseudoR (and asking participants to choose between them).
Version B was distributed in two variants (B1 and B2) with different (and completely disjoint) sets of pseudowords and image generations.

For all surveys, we selected images by using the image closest to the centroid in CLIP embedding space among images in a 50-item minibatch of generations (as described in Section \ref{sec:imgen}), in order to reduce variance due to the stochastic nature of the image generation process.

In version B of the survey, each question displayed one image generated by a pseudoword from \PseudoS and another generated by a pseudoword from \PseudoR, as seen in the right-hand screenshot in Figure \ref{fig:supp_survey}. No pseudoword appeared in more than one question. We randomly selected pseudowords for use in these questions, but in order to focus on the presence of sound symbolism and avoid participants easily identifying real objects, we manually removed pseudowords close to real English words, as discussed in Section \ref{sec:eng_sim}.

\subsection{Results Analysis}

175 people participated in our survey: 100 saw version A, 50 saw version B1, and 25 saw version B2. Of these, 87 reported that they had not heard of the \emph{kiki--bouba} effect before taking the survey (47 for A, 30 for B1, 10 for B2); we take this into account below and show that our results have similar implications whether including these participants or not.

Overall, participants answered correctly more often than the 50\% baseline accuracy expected from random guessing: They answered with 73\% accuracy in the \emph{kiki--bouba} setting (version A) and 55\% in the general pseudoword setting (versions B1 and B2). When only considering participants who reported that they had not heard of the effect before, the accuracy values were 70\% and 56\% respectively.

To determine statistical significance and the overall effect size, we control for inter-subject and inter-item variation with a mixed-effects logistic regression model, using the \texttt{Lmer} implementation in the \texttt{pymer4} library~\cite{pymer4}. We regress whether a question is answered correctly by a respondent (categorical response variable: 0 for incorrect answer, 1 for correct answer), treating question identity as a fixed effect and respondent identity as a random effect. We perform this separately for setting A and setting B; in the former, each subject answered all 20 questions; in the latter, each subject answered either the 20 questions of version B1 or the 20 questions of version B2, 40 distinct possible question categories total.

For setting A (\emph{kiki} vs. \emph{bouba}), our model results in an intercept estimate corresponding to a 89\% overall success probability ($p < 0.001$) when calculated over all 100 respondents. When using only the 47 who reported not having heard of the effect, the corresponding intercept estimate is 86\% ($p < 0.001$). For setting B (random pseudowords), our model results in an intercept estimate corresponding to a 78\% overall success probability ($p < 0.001$) when calculated over all 75 respondents. When using only the 40 who reported not having heard of the effect, the corresponding intercept estimate is 74\% ($p < 0.015$).

Overall, we find that the sound symbolism exhibited by our text-to-image model significantly accords with human sound symbolic associations; these results are reflected whether or not we restrict our analysis to subjects who had not heard of the \emph{kiki--bouba} effect before.

\end{document}